\documentclass[10pt,twocolumn,letterpaper]{article}

\usepackage{cvpr}              

\usepackage{graphicx}
\usepackage{amsmath}
\usepackage{amssymb}
\usepackage{booktabs}
\usepackage{pifont}

\usepackage{bbding}
\usepackage{soul}
\usepackage{bibunits}
\usepackage{bm}
\usepackage{array}
\usepackage{pdflscape}
\usepackage{makecell}
\usepackage{framed,multirow}
\usepackage{threeparttable}
\usepackage{amssymb}
\usepackage{pifont}
\usepackage{listings}
\usepackage{pythonhighlight}
\usepackage{float}
\usepackage{mathtools}
\usepackage{cuted}

\usepackage{algorithmicx}
\usepackage{algorithm}
\usepackage{algpseudocode}

\usepackage{multicol}
\usepackage{multirow}
\usepackage{pifont}

\newcolumntype{P}[1]{>{\centering\arraybackslash}p{#1}}
\newlength\savewidth
\def\arrvline{\hfil\kern\arraycolsep\vline\kern-\arraycolsep\hfilneg}
\definecolor{mygray}{gray}{.9}
\definecolor{Highlight}{HTML}{39b54a}  

\newcolumntype{x}[1]{>{\centering\arraybackslash}p{#1pt}}
\newcolumntype{z}[1]{>{\raggedright\arraybackslash}p{#1pt}}

\definecolor{citecolor}{HTML}{0071BC}
\definecolor{linkcolor}{HTML}{ED1C24}

\usepackage[T1]{fontenc}

\newcommand{\ourdataset}{SceneBench}  %
\newcommand{\ourbench}{SceneBench}  

\newcommand{\ourrag}{{Scene-RAG}}

\usepackage[T1]{fontenc}\usepackage[utf8]{inputenc}\usepackage{textcomp}

\usepackage[table,dvipsnames]{xcolor}
\usepackage{booktabs}
\usepackage{colortbl}      

\usepackage{booktabs}
\usepackage{threeparttable}
\usepackage{multirow}
\usepackage{makecell}
\usepackage{pifont}    
\usepackage{siunitx}
\newcommand{\best}[1]{\textbf{#1}}

\newcolumntype{C}{>{\centering\arraybackslash}m{1.3cm}}
\newcolumntype{Y}{S[table-format=2.1]}
\newcommand{\model}[1]{\textsc{#1}} 

\definecolor{OpenSrcBlue}{rgb}{0.921,0.961,1.0}  
\definecolor{headerpurple}{RGB}{230,238,250}
\newcommand{\proprow}{\rowcolor{gray!11}}
\newcommand{\openrow}{\rowcolor{OpenSrcBlue}}

\definecolor{cvprblue}{rgb}{0.21,0.49,0.74}
\usepackage[pagebackref,breaklinks,colorlinks,allcolors=cvprblue]{hyperref}

\def\confName{CVPR}
\def\confYear{2026}

\title{
Seeing the Scene Matters: Revealing Forgetting \\in Video Understanding Models with a Scene-Aware Long-Video Benchmark
}

\author{
\renewcommand{\arraystretch}{1.12}
\begin{tabular}{cccc}
\parbox[t]{0.23\linewidth}{\centering
Seng Nam Chen\thanks{Equal contribution. Contact: hc666@cam.ac.uk.}\\
\small\mbox{CUHK (SZ)}
}
&
\parbox[t]{0.23\linewidth}{\centering
Hao Chen$^{*}$\\
\small\mbox{University of Cambridge}
}
&
\parbox[t]{0.23\linewidth}{\centering
Chenglam Ho\\
\small\mbox{UESTC}
}
&
\parbox[t]{0.23\linewidth}{\centering
Xinyu Mao\\
\small\mbox{CUHK}
}
\end{tabular}
\\[1.3em]
\begin{tabular}{ccc}
\parbox[t]{0.31\linewidth}{\centering
Jinping Wang\\
\small\mbox{GPNU}
}
&
\parbox[t]{0.31\linewidth}{\centering
Yu Zhang\\
\small\mbox{Shanghai Jiao Tong University}
}
&
\parbox[t]{0.31\linewidth}{\centering
Chao Li\thanks{Corresponding author.}\\
\small\mbox{University of Cambridge}
}
\end{tabular}\\[1.0em]
{\small \texttt{Website:} \textcolor{pink}{\url{https://huggingface.co/datasets/SinerChen/SceneBench}}}
}

\begin{document}
\maketitle


\begin{abstract}

Long video understanding (LVU) remains a core challenge in multimodal learning. Although recent vision-language models (VLMs) have made notable progress, existing benchmarks mainly focus on either fine-grained perception or coarse summarization, offering limited insight into temporal understanding over long contexts. In this work, we define a scene as a coherent segment of a video in which both visual and semantic contexts remain consistent, aligning with human perception. This leads us to a key question: \textit{can current VLMs reason effectively over long, scene-level  contexts?} To answer this, we introduce a new benchmark, \ourdataset, designed to provide scene-level challenges. Our evaluation reveals a sharp drop in accuracy when VLMs attempt to answer scene-level questions, indicating significant forgetting of long-range context. 
To further validate these findings, we propose Scene Retrieval-Augmented Generation (Scene-RAG), which constructs a dynamic scene memory by retrieving and integrating relevant context across scenes. This Scene-RAG improves VLM performance by  {+2.50\%}, confirming that current models still struggle with long-context retention. We hope \ourdataset~ will encourage future research toward VLMs with more robust, human-like video comprehension.

\end{abstract}

\section{Introduction}

Humans do not watch videos frame by frame; we follow stories as they unfold~\cite{hasson2008neurocinematics,smith2012attentional}. Each story is structured through scenes, which serve as the basic units of narrative understanding~\cite{zacks2001event}. Our minds naturally organize continuous visual input into coherent segments where appearances and meanings remain consistent~\cite{zacks2007event}. For example, in an episode of \textit{Friends}, “Monica’s apartment” or “Central Perk” each forms a distinct scene that recurs throughout the narrative. Within these scenes, we remember who appeared earlier, why an action happened, and how one moment leads to the next~\cite{trabasso1982causal}. When a scene changes, our mental model resets; when it returns, we recall what came before~\cite{zacks2007event}. This scene-based organization helps humans maintain continuity over time and make sense of long narratives~\cite{mariola2022event}.

\begin{figure}
    \centering
    \includegraphics[width=1\linewidth]{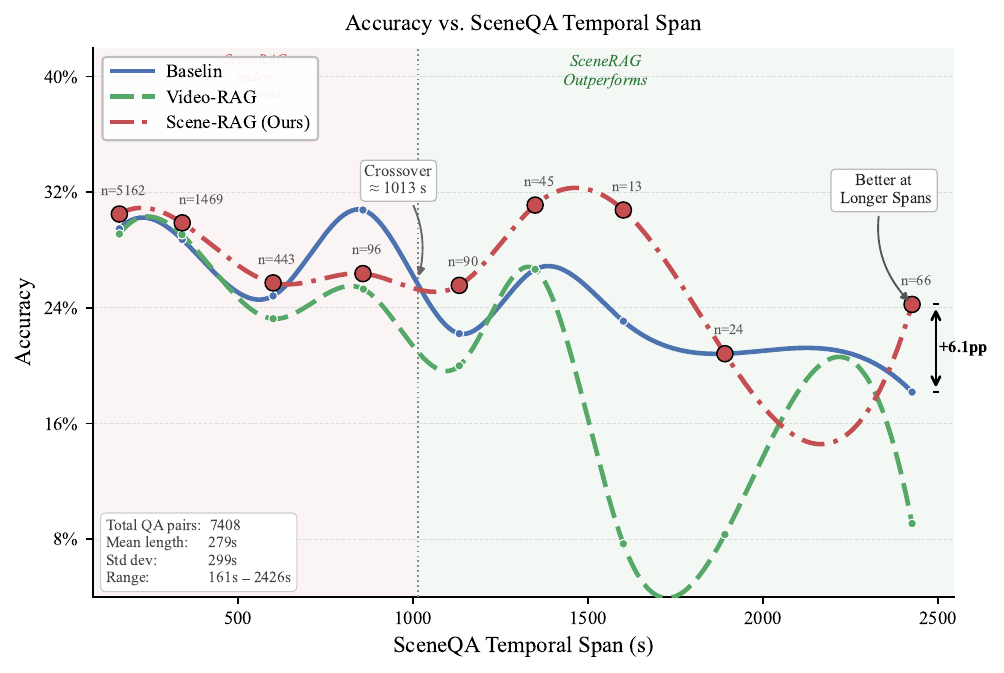}
  \caption{Longer SceneQA distances lead to lower accuracy. Video-RAG~\cite{luo2024video} improves the baseline in short- and long-range settings but struggles in the mid-range, while our  \ourrag~achieves consistent gains, especially for mid- and long-term reasoning. Curves are smoothed using a rolling mean. Results use Qwen2.5-VL~\cite{qwen2.5-VL}. Solid markers denote actual measurements; curves are cubic-spline interpolations for visual clarity. }
    \label{fig:teaser}
\end{figure}

\begin{figure*}[t]
    \centering
    \includegraphics[width=1\linewidth]{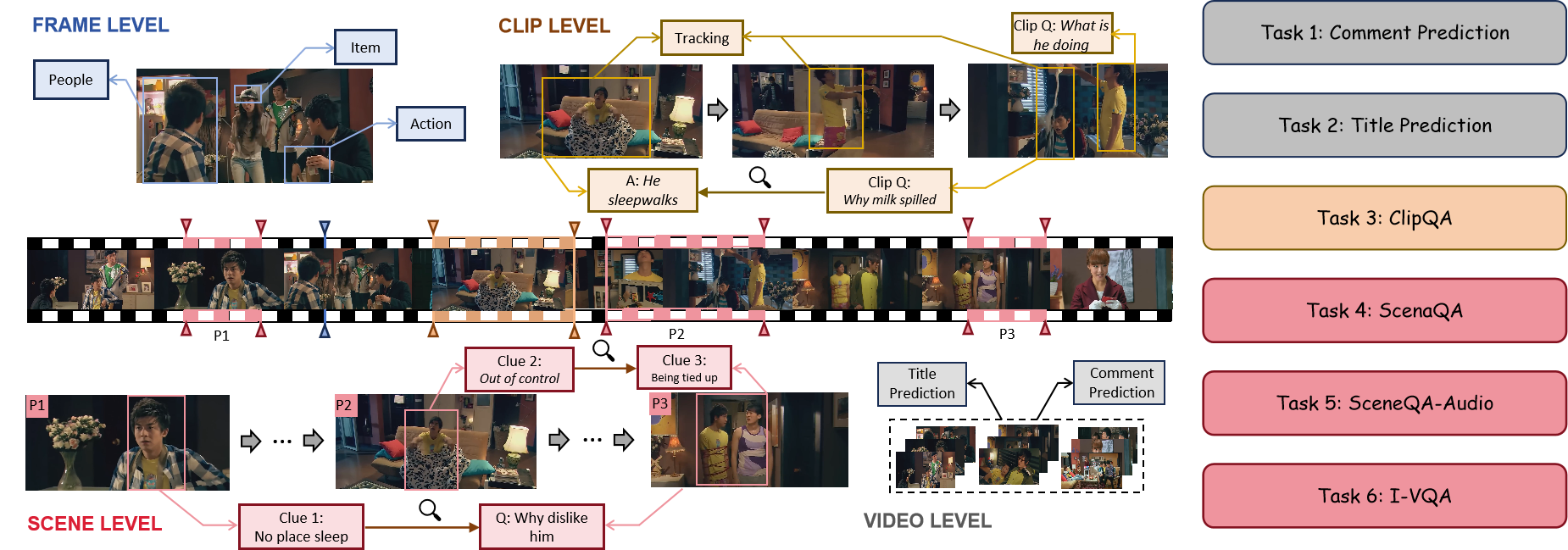}
    \caption{We divide video information into frame, clip, scene, and video levels. For frame-level information, it focuses mainly on describing the details of the subject within the frame. Clip-level information includes temporal information but can only describe the objective behavior of objects. Scene-level information contains a lot of clip-level information and forms a complete scenario event, while video consists of a lot of scene-level content, with related scenes forming a logical storyline based on cause and effect. For convenience of quantification, we use two minutes to distinguish between clip and scene levels.}
    \label{fig:task_challenges}
\end{figure*}

 Recent advances in multi-modal large language models (MLLMs) have significantly enhanced their ability to integrate and interpret text, images, and videos~\cite{liu2023visual, chen2024internvl, guo2024mammoth, zhang2024flash, zhu2025internvl3, team2025kimi}. These models demonstrate impressive performance on short clips, \eg~recognizing actions, describing events, and answering visual questions over spans of a few seconds~\cite{fu2023mme, mvbench2023, liu2023mmbench, yue2023mmmu}. However, as videos grow longer, such as movies, sports broadcasts, or documentaries, the models begin to lose coherence. Earlier information fades and overall narrative understanding deteriorates~\cite{moviechat2023, zhang2024longva, li2024videovista}. As shown in Figure~\ref{fig:teaser}, the longer the QA distance (the gap between a question and its answer), the worse the model's response quality becomes.

Existing long-video understanding benchmarks \cite{fu2024videommefirstevercomprehensiveevaluation, ma2025iv, li2024videovista, zhou2024mlvu, 2024arXiv240509711W,chen2024360} only partially address this challenge. Most existing benchmarks rely on short or isolated video clips~\cite{msrvtt2016, seedbench2023, cvrr-2024}, or they evaluate different temporal scales on entirely different videos~\cite{zhou2024mlvu}. Consequently, these benchmarks conflate temporal granularity with content variability, making it difficult to disentangle whether a model truly understands temporal structure or merely adapts to changing content. What remains missing is a rigorous, \textit{within-video} evaluation framework that examines how models reason across scenes.

Motivated by this gap, we introduce \ourdataset, a benchmark for evaluating long-context understanding through the lens of scenes in video. It comprises three core challenges: 
(1) Scene Question Answering (SceneQA), which tests whether a model can aggregate dispersed visual clues within a scene to answer complex questions, along with SceneQA-Audio, an extension that incorporates audio cues such as dialogue and ambient sounds; 
(2) Inverse Visual QA (I-VQA), which reverses the reasoning direction that given an answer, the model must infer the underlying question, testing its grasp of causal and contextual links; and (3) Comment Prediction (CP), which measures a model’s ability to generate human-like narrative responses that depend on long-term story comprehension. Finally, to capture understanding across multiple temporal scales, \ourdataset\ also includes ClipQA and title prediction (TP), assessing the transition from fine-grained scene reasoning to holistic video-level comprehension.

Across our benchmark, a clear pattern emerges. Existing MLLMs excel on short, self-contained clips (ClipQA) but their performance collapses on scene-level tasks (SceneQA, i-VQA, SceneQA-Audio, and CP). The longer the temporal dependency, the steeper the decline. This reveals a key weakness: current models struggle to retain and retrieve long-term contextual information needed for coherent story understanding. To further investigate this phenomenon, we introduce scene retrieval-augmented generation (\ourrag), which equips models with an external, dynamic memory of scenes, both visual and auditory. Experiments show that \ourrag~improves performance across our benchmark and other long video understanding (LVU) datasets, reinforcing our finding that MLLMs suffer from long-term memory forgetting. Retrieval augmentation thus also proves to be an effective strategy for mitigating long-context forgetting. 

\newcommand{\cmark}{\textcolor{green!60!black}{\ding{51}}} 
\newcommand{\xmark}{\textcolor{red!70!black}{\ding{55}}}   

\begin{table*}[t]
\centering
\caption{Comparison of SceneBench with existing long-video understanding benchmarks. SceneBench offers a balanced number of videos, diverse genres, and high–time-range, scene-based QA annotations. The average temporal distance between each question and its corresponding answer in SceneQA is 262s, with a standard deviation of 310s, illustrating the wide temporal diversity of the dataset. ``$\dagger$'' denotes automatically or partially automatically generated QA pairs.}
\label{tab:comparison}
\resizebox{1.0\textwidth}{!}{
\begin{tabular}{l
c
c
c c c c c c
c c c}
\toprule
\textbf{Benchmark} &
\textbf{Year} &
\makecell{\textbf{\#}\textbf{Videos}} &
\makecell{\textbf{Len.}\textbf{(s)}} &
\makecell{\textbf{\#}\textbf{Tasks}} &
\makecell{\textbf{\#}\textbf{Genres}} &
\makecell{\textbf{\#QA}\\\textbf{Pairs}} &
\makecell{\textbf{QA}\\\textbf{Len. (s)}} &
\makecell{\textbf{QA}\\\textbf{Density}} &
\makecell{\textbf{Multi-}\\\textbf{level}} &
\makecell{\textbf{Scene}\\\textbf{Reliance}} &
\makecell{\textbf{Audio}\\\textbf{Reliance}} \\
\midrule
LVBench~\cite{wang2025lvbench}              & 2025 & 103  & 4,101 & 6  & 21 & 1,549            & --  & 0.38 & \xmark & \xmark & \xmark \\
LongVideoBench~\cite{wu2024longvideobench}  & 2024 & 3,763 & 473   & 17 & 10 & 6,678            & --  & 14.1 & \xmark & \xmark & \xmark \\
HourVideo~\cite{chandrasegaran2024hourvideo} & 2024 & 500  & 2,742 & 4  & 16 & 12,976$^\dagger$ & --  & 4.7  & \xmark & \xmark & \xmark \\
ALLVB~\cite{tan2025allvb}                    & 2025 & 1,376 & 7,200 & 9  & 30 & 252,000$^\dagger$ & --  & 35.0 & \xmark & \xmark & \xmark \\
Video-MME~\cite{fu2025video}                 & 2025 & 900  & 1,024 & 12 & 30 & 2,700            & --  & 2.6  & \cmark & \xmark & \xmark \\
MoVQA~\cite{zhang2023movqa}                  & 2023 & 930  & 992   & 6  & 1  & 21,953           & --  & 22.1 & \cmark & \xmark & \xmark \\
MLVU~\cite{zhou2024mlvu}                     & 2024 & 1,730 & 930   & 9  & 31 & 3,102            & --  & 3.3  & \cmark & \xmark & \xmark \\
ScaleLong~\cite{ma2025scalelong}             & 2025 & 269  & 5,160 & 5  & 36 & 1,747            & --  & 0.34 & \cmark & \xmark & \xmark \\
\midrule
\textbf{\ourdataset~(Ours)}                  & \textbf{2025} & \textbf{2,485} & \textbf{1,978} & \textbf{6} & \textbf{6} & \textbf{8,903} & \textbf{262} & \textbf{4.50} & \textbf{\cmark} & \textbf{\cmark} & \textbf{\cmark} \\
\bottomrule
\end{tabular}
}
\end{table*}

To summarize, our contributions are threefold: 
\begin{enumerate}
    \item We present \ourdataset, the first benchmark that systematically evaluates scene-level understanding in long videos, comprising \textit{2,485} videos (avg. \textit{1,978} seconds) and \textit{8,903} question–answer pairs across 6 challenge types. It systematically evaluates models' long-context reasoning capabilities, preventing them from exploiting simple plot matching.
    
    \item We conduct extensive evaluations showing that existing MLLMs struggle with scene-based reasoning, with performance drops sharply as scenes become longer. 
    
    \item To mitigate this long-term forgetting, we introduce Scene-RAG, which enhances models with retrieval-augmented memory and improves their ability to reason over long temporal contexts.  
\end{enumerate}

\section{Related Work}

\smallskip\noindent\textbf{MLLMs for Video Understanding.} 
Recent works further adapt Multimodal large language models (MLLMs) to videos using temporal adapters or video-instruction datasets~\cite{videollama, videochatgpt2023, mplug-owl-2023, videollava2023}.
However, most are limited to short clips due to computational and contextual constraints. Several approaches attempt to extend context length or compress visual tokens. For example, LLaMA-Vid~\cite{llama-vid2023} compresses each frame into a few tokens, while MovieChat~\cite{moviechat2023} and MA-LMM~\cite{malmm2024} introduce memory modules for recursive reasoning. Long-video methods such as LWM~\cite{liu2024world}, LongVA~\cite{zhang2024longva}, and Video-XL~\cite{shu2024videoxl} enlarge temporal context, and retrieval-based models~\cite{R-VLM-2023, videoagent2024} selectively sample key frames.
Despite these efforts, long-form comprehension, especially over multi-hour videos with complex narratives, remains an open challenge.


\smallskip\noindent\textbf{Video Understanding Benchmarks.} 
Early video benchmarks mainly evaluate short clips for captioning, action recognition, or reasoning tasks~\cite{activitynetqa2019, star2021, msrvtt2016, howto100m2019}.
MVBench~\cite{mvbench2023} provides a unified short-video QA benchmark, but it still lacks coverage of long-term temporal understanding. 
To evaluate long videos, datasets such as LLaMA-Vid~\cite{llama-vid2023}, MovieChat~\cite{moviechat2023}, and MoVQA~\cite{zhang2023movqa} utilize movies or egocentric videos to design question–answer tasks. While valuable, these benchmarks often depend on manual annotations or timestamp-specific questions, leading to fragmented and inconsistent evaluations.
Recent works (\eg MLVU~\cite{zhou2024mlvu}, Video-MME~\cite{fu2025video}, LVBench~\cite{wang2025lvbench}) attempt broader coverage across perception, reasoning, and summarization, yet still lack a unified and scalable framework for constructing long-video tasks.


\begin{figure*}
    \centering
    \includegraphics[width=1\linewidth]{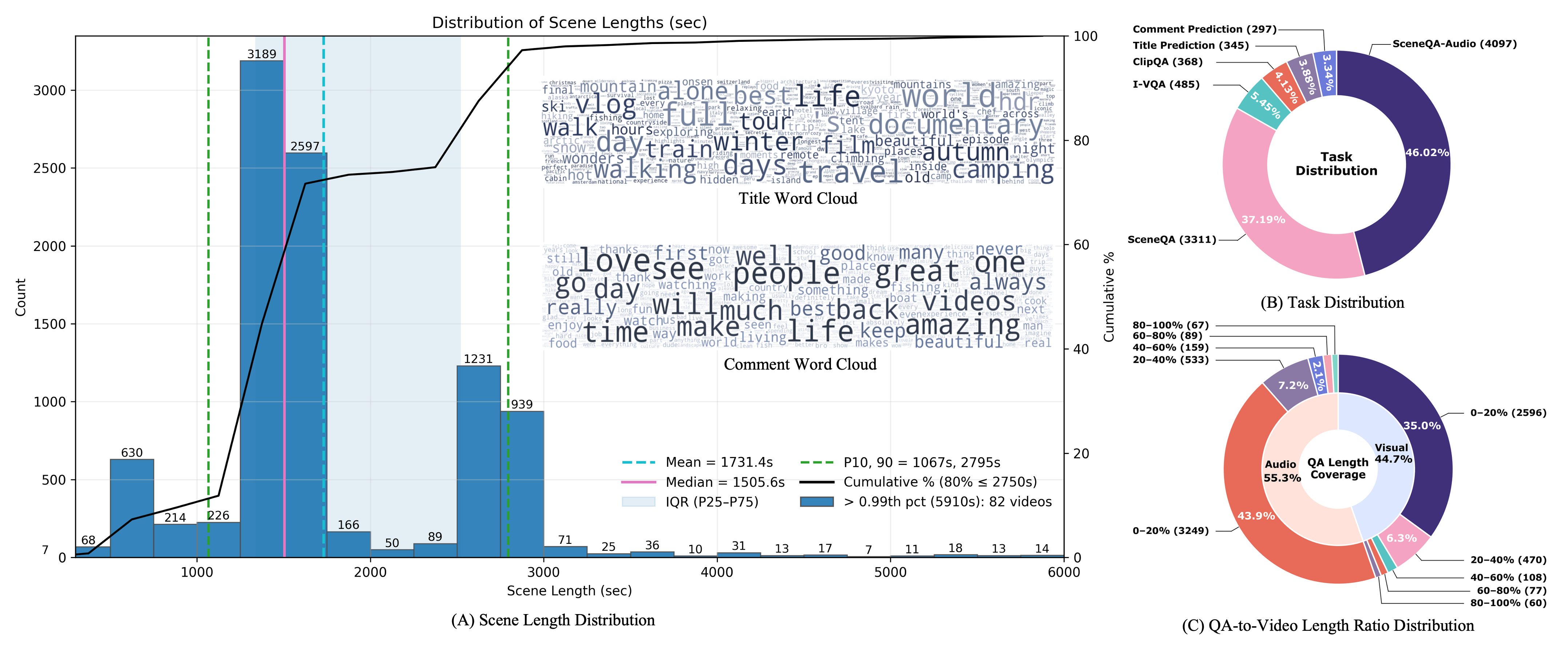}
    \caption{Statistical overview of our \ourbench~benchmark. 
    (A) Distribution of scene lengths. 
    (B) Distribution of task counts. 
    (C) Distribution of  SceneQA and SceneQA-Audio length duration proportion over the full video.  }
    \label{fig:overview_stat}
    \vspace{-0.2cm}
\end{figure*}

\section{\ourdataset: Design and Construction}

In this section, we first provide an overview of \ourdataset, outlining its components and task challenges. We then describe the structure of each task, followed by a discussion of the real-world issues encountered during task construction. Finally, we present detailed statistics that illustrate the characteristics of the dataset in Figure~\ref{fig:overview_stat}.

\subsection{Overview}

The \textit{\ourdataset}~long video understanding benchmark consists of 2,485 videos with an average duration of 1,978 seconds ($\sim$33 minutes), ranging from as short as 1 minute to over 4 hours. It includes 8,903 QA pairs across six task categories: 3,311 SceneQA, 4,097 SceneQA-Audio, 368 ClipQA, 485 I-VQA, 345 Title Prediction, and 297 Comment Prediction. Table~\ref{tab:comparison} presents a  statistical comparison between \ourdataset~and existing long-video understanding benchmarks.

\subsection{Task Design Principles}

We design six tasks in total: (1) Scene Question Answering (SceneQA), (2) its audio-based extension (SceneQA-Audio), (3) Clip Question Answering (ClipQA), (4) Inverse Video Question Answering (I-VQA), (5) Comment Prediction, and (6) Title Prediction.

\smallskip\noindent\textbf{SceneQA and Its Counterparts.} 
Previous definitions of event-level or scene-level problems typically focused solely on the temporal span between cues (e.g., the number of seconds between two key frames), while neglecting the duration of the cues themselves. This “spanning distance”-based approach fails to adequately capture the complexity of events within long videos, as many key events are not instantaneous occurrences but rather consist of dialogues, actions, or interactions spanning several minutes.
To more clearly delineate the objectives of long-form video understanding, we provide a stricter definition for scene-level problems:
The effective cues required to answer this question must span several minutes of video.
This specification ensures that scene-level problems genuinely require models to integrate long-range, multi-segment, non-instantaneous cues rather than relying on short local segments for resolution. This better aligns with the semantic structure of complex events in real-world long-form videos. 

For \textbf{SceneQA} construction, we deliberately ensure that clues for SceneQA span at least two minutes and provide as much detail as possible about the clues, thereby filtering out transient or one-off moments that lack narrative continuity. During question design, we avoid object-centric or attribute-level queries and instead focus on plot-specific reasoning questions that require deep contextual understanding rather than pattern matching. Many questions and answers extend across different temporal or spatial contexts, compelling models to integrate information from multiple points in time to infer the correct response.

\textbf{SceneQA-Audio} further extends this concept by introducing the auditory modality as an essential reasoning signal. In real-world storytelling, dialogues, ambient sounds, and background music often carry critical cues that cannot be inferred visually. Inspired by this, SceneQA-Audio builds upon the SceneQA setting but requires models to extract and reason over plot-relevant information embedded in the audio track, offering complementary insights that enhance narrative understanding beyond the visual only.

In addition, we introduce \textbf{ClipQA} to complement SceneQA by providing QA pairs at shorter temporal scales. Each clip clue in our dataset spans around 30 seconds, capturing concise narrative units within the broader storyline. 



\smallskip\noindent\textbf{I-VQA: Inverse Understanding.} 
Traditional video QA presents a clip and asks for an answer, which often allows models to rely on linguistic cues or short-term visual patterns without true temporal reasoning. Motivated by this, we propose \textbf{Inverse Video Question Answering (I-VQA)}, where the answer is given and the model must identify the correct question based on the entire video. This inversion forces reasoning from effect to cause, requiring the model to search across time to locate the supporting visual evidence and reconstruct the narrative context in which the answer is valid. Since related events may be temporally distant or separated by irrelevant scenes, I-VQA demands global temporal tracking, causal inference, and memory-based reasoning. Through this design, I-VQA explicitly enforces long-video understanding beyond surface-level pattern matching.

\smallskip\noindent\textbf{Comment and Title Prediction.} 
We further propose two  tasks to evaluate high-level understanding. \textbf{Comment Prediction} (CP) requires the model to determine whether the comment pertains to this specific video, which often reflects emotional reactions, thematic interpretation, or implicit reasoning about prior and future events. This task requires the model not only to correctly understand visual cues displayed in long videos, but also to comprehend the implicit semantic relationship between comments and videos. In contrast, \textbf{Title Prediction} (TP) focuses on summarizing the central theme or topic of a video in a concise phrase. Since titles typically reflect high-level or static information rather than detailed temporal reasoning, this task depends less on long-term context and more on global semantic abstraction.

\begin{figure*}[tp]
    \centering
    \includegraphics[width=1\linewidth]{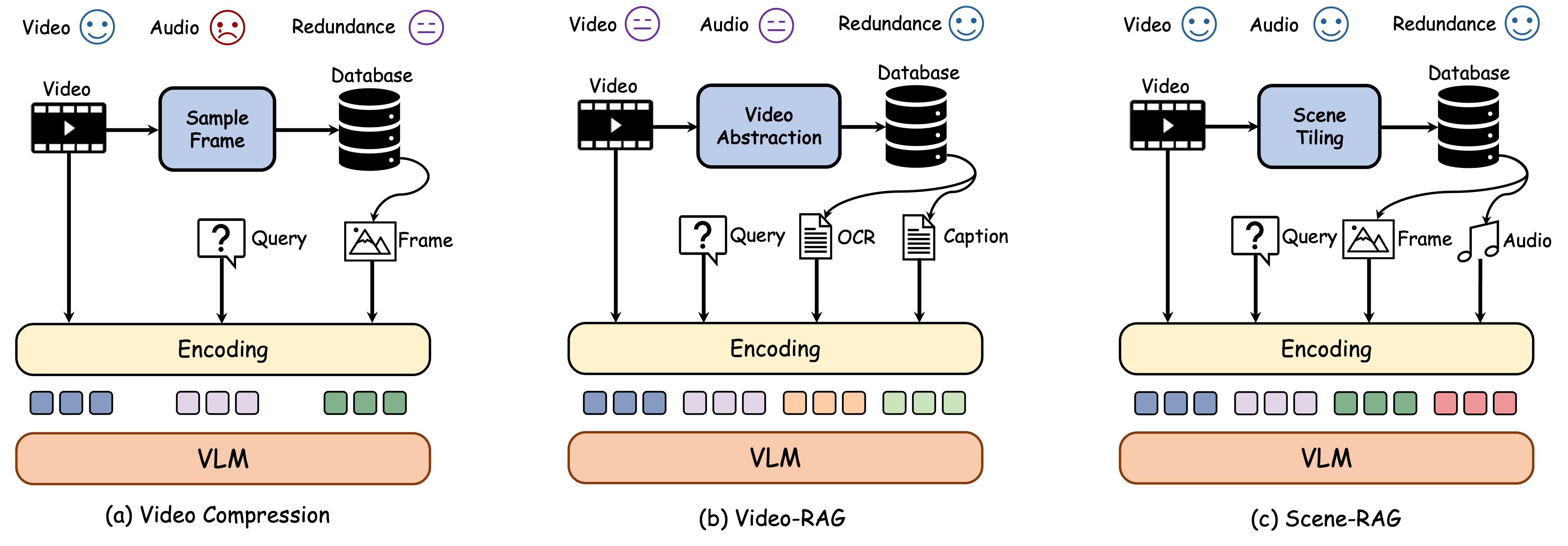}
    \caption{Comparison of Scene-RAG with traditional RAGs working on video understanding tasks. Scene-RAG first aggregates long-range visual scenes via scene tiling, stores aligned visual–audio evidence, and retrieves task-relevant segments conditioned on user queries.}
    \label{fig:scene-rag}
\end{figure*}

\subsection{Collection and Labeling Process}

\smallskip\noindent\textbf{Collection.} We collect all videos in our benchmark from publicly available sources on YouTube under conditions that permit redistribution. To ensure content diversity and narrative completeness, we select videos from a wide range of genres, including films, vlogs, and documentaries. All videos are manually verified to exclude, age-restricted or low-quality content, ensuring that every video in the dataset can be legally used for research purposes.

\smallskip\noindent\textbf{Labeling. } The annotation process follows a carefully designed multi-stage pipeline to ensure quality and consistency. \textit{Comment} labeling begins with annotators selecting remarks that reflect genuine viewer engagement, prioritizing those that are widely appreciated and highly rated. Additionally, we explicitly require annotators to avoid selecting generic comments applicable to any video (e.g., “do a good job”). Such generalized expressions lacking content dependency fail to capture the unique narrative or plot of a specific video, thereby rendering them unsuitable for evaluating the model's semantic understanding of particular videos. The final comment annotations are determined through consensus among multiple annotators. 

For the \textit{SceneQA pairs}, annotators first divide each video into scenes according to narrative  coherence. These initial segments are reviewed by another annotator to verify the accuracy of temporal boundaries and maintain consistency. Based on the finalized segments, annotators create question and answer pairs and other related annotations, including the time required to locate each answer within the video, following detailed task-specific guidelines. Each annotation is independently validated by at least another reviewer to ensure factual correctness.

For the \textit{I-VQA} task, annotators enforce strict consistency among all candidate questions. Specifically, when the correct question belongs to a particular reasoning category (e.g., causal reasoning), all distractor questions are constructed within the same category, avoiding mismatches such as mixing “causal” and “how-to” question types. Moreover, each candidate question is designed to be semantically compatible with the provided answer, ensuring that none of the distractors can be eliminated purely through linguistic analysis. This prevents models from exploiting language priors and forces them to ground their decisions on actual video evidence.

It is noteworthy that as the QA time range extends from minute-scale windows to the much longer, multi-minute spans required for SceneQA, the annotation process becomes more challenging and often ambiguous. We discuss the specific issues we encountered and our corresponding solutions in the supplementary material.



\section{Scene-RAG: Long Context Retrieval}
MLLMs struggle when videos extend beyond a few minutes: visual details are lost due to often sparse frame sampling, or long-term memory loss, and audio cues are often ignored altogether. Retrieval-Augmented Generation (RAG) offers a natural solution, by providing models with external, query-relevant memory.  The discussion of recent efforts on RAG for long video~\cite{jeong2025videoragretrievalaugmentedgenerationvideo,luo2024videoragvisuallyalignedretrievalaugmentedlong,zhang2025qframequeryawareframeselection,yuan2025memoryenhancedretrievalaugmentationlong} are provided in the supplementary material.

\subsection{Scene-RAG}

We introduce Scene-RAG, a retrieval-augmented framework designed for scene-based long-term reasoning,  as illustrated in Figure~\ref{fig:scene-rag}. 
Instead of treating a video as a sequence of uniform clips, Scene-RAG organizes its memory around semantic scenes, continuous segments that encapsulate coherent visual and audio narratives.

Scene-RAG consists of three stages: (1) Scene Tiling to detect and aggregate semantically coherent scene segments; (2) Memory Construction to encode multimodal scene representations for retrieval; (3) Query Retrieval to decompose a query and retrieve relevant scene memories for reasoning.

\smallskip\noindent\textbf{Scene Tiling.}
Conventional frame sampling methods often produce redundant or disjoint frames that disrupt temporal reasoning. To form coherent visual units, we aggregate frames into continuous scene segments through \textit{Total Variation with L1 regularization} (TV-L1). Let $s_1,\dots,s_n$ denote a similarity sequence along the video timeline. TV-L1 minimizes:
\begin{equation}
  \min_{x\in\mathbb{R}^n}
  \;\frac{1}{2}\sum_{t=1}^n (x_t - s_t)^2
  \;+\;
  \lambda \sum_{t=2}^n |x_t - x_{t-1}|,
\end{equation}
\noindent 
where $x_t$ is the denoised similarity at time $t$, $s_t$ is the raw similarity value, and $\lambda>0$ controls the strength of the temporal regularization. The quadratic fidelity term encourages $x_t$ to stay close to $s_t$, whereas the L1 total-variation term penalizes abrupt temporal changes, yielding a piecewise-constant approximation whose plateaus correspond to semantically coherent intervals.   

To identify salient segments, we compute a statistical threshold $k=\mu_x+\alpha\sigma_x$, where $\mu_x$ and $\sigma_x$ are the mean and standard deviation of the denoised sequence and $\alpha$ is a tunable sensitivity parameter. Consecutive time indices with $x_t\ge k$ form high-activation runs that indicate meaningful scenes. Any detected segment whose duration falls below a minimum acceptable length $L_{\min}$ is discarded to eliminate spurious or noisy intervals.


\smallskip\noindent\textbf{Multimodal Memory Construction}

The extracted scene segments are encoded using InternVideo2~\cite{wang2024internvideo2} to capture visual semantics. Each segment is processed as an independent video stream to obtain high-level embeddings that respect temporal consistency. In parallel, the corresponding audio tracks are transcribed and captioned using Qwen-Audio2~\cite{chu2024qwen2}, producing concise textual descriptions that summarize speech, ambient sound, and background music. Visual and audio representations are aligned along the timeline to form multimodal scene entries. The resulting collection of scene-level embeddings and captions constitutes the memory bank for retrieval.

\smallskip\noindent\textbf{Query Retrieval}

At inference, the model receives a user query that may reference events, objects, or relationships spanning multiple scenes. We first use Qwen3 14B~\cite{yang2025qwen3} to decompose the query into fine-grained textual clues, which serve as semantic anchors for retrieval. Each clue is encoded into the same feature space as the multimodal memory, enabling similarity search over the scene embeddings. The top retrieved scenes are then fused and provided to the MLLM for reasoning and generation. This ensures the model grounds its answers in contextually relevant, temporally aligned evidence, mitigating long-term memory decay.

\begin{table*}[t]
\centering
\caption{\textbf{Benchmark results on the \ourdataset{}.} ``Frame Count''  indicates the number of frames used as input. Comm.: Comment.}
\label{tab:mllm_2024plus}
\setlength{\tabcolsep}{4.8pt}
\renewcommand\arraystretch{1.12}

\resizebox{0.95\textwidth}{!}{
\begin{tabular}{l c|c| c c c c c c|c}
\toprule

\rowcolor{headerpurple}
\textbf{Method} &
\textbf{Release} &
\textbf{Frame Count} &
\makecell{\textbf{Title}\\\textbf{Pred.}} &
\makecell{\textbf{Comm.}\\\textbf{Pred.}} &
\makecell{\textbf{Clip}\\\textbf{QA}} &
\makecell{\textbf{Scene}\\\textbf{QA}} &
\makecell{\textbf{SceneQA}\\\textbf{-Audio}} &
\makecell{\textbf{I-VQA}} &
\makecell{\textbf{Avg.}} \\

\midrule

\multicolumn{10}{c}{\textit{Proprietary MLLMs}} \\
\midrule

\proprow \model{Gemini 2.5 Pro}~\cite{comanici2025gemini25pushingfrontier} & 2025 & 1fps  & 99.4 & 92.9 & 80.5 & 60.7 & 65.8 & 82.4 & \best{80.3} \\
\proprow \model{Gemini 2.5 Flash-Lite}~\cite{comanici2025gemini25pushingfrontier} & 2025 & 1fps  & 99.1 & 78.2 & 48.4 & 25.6 & 30.1 & 41.6 & 53.8 \\
\proprow \model{Kimi 2.5v}~\cite{kimiteam2026kimik25visualagentic} & 2025 & 128  & 99.4 & 90.9 & 73.1 & 54.0 & 59.6 & 60.7 & 73.0 \\
\midrule

\multicolumn{10}{c}{\textit{Open-source MLLMs}} \\
\midrule

 \openrow\model{MovieChat}~\cite{moviechat}               & 2023 & 2048 & 25.2 & 28.6 & 52.2 & 21.9 & 21.4 & 21.6 & 28.5 \\

\model{VideoChat2}~\cite{li2023mvbench}                   & 2023 & 16   & 98.3 & 72.5 & 45.4 & 31.5 & 36.1 & 33.2 & 52.8 \\

\openrow\model{Video\mbox{-}CCAM}~\cite{videoccam}                & 2023 & 96   & 98.8 & 79.7 & 51.0 & 29.3 & 33.3 & 37.0 & 54.9 \\

\model{LLaMA\mbox{-}VID}~\cite{llama-vid2023}             & 2023 & 1fps & 94.5 & 56.1 & 66.3 & 23.7 & 24.3 & 32.0 & 49.5 \\

\openrow \model{TimeChat}~\cite{timechat}                 & 2024 & 96   & 17.7 & 18.3 & 19.3 & 23.9 & 26.1 & 7.3  & 18.8 \\

\model{LongVA}~\cite{longva}                     & 2024 & 128  & 96.8 & 71.1 & 54.3 & 27.7 & 29.5 & 40.1 & 53.3 \\
\openrow \model{Long\mbox{-}LLaVA}~\cite{longva}                   & 2024 & 64   & 96.2 & 79.7 & 57.1 & 26.1 & 25.4 & 33.0 & 52.9 \\
 
 \model{MA\mbox{-}LMM}~\cite{malmm2024}           & 2024 & 1000 & 69.9 & 46.3 & 38.5 & 27.7 & 32.7 & 30.8 & 41.0 \\

\openrow \model{mPLUG\mbox{-}Owl3\mbox{-}V}~\cite{ye2024mplug}  
                                                        & 2024 & 64   & 91.3 & 74.1 & 54.1 & 24.3 & 28.7 & 40.7 & 52.2 \\
\model{InternVL2.5\mbox{-}7B}~\cite{chen2024expanding}  
                                                        & 2024 & 16   & 98.6 & 79.3 & 48.3 & 27.9 & 30.5 & 40.7 & 54.2 \\
\openrow \model{LLaVA\mbox{-}OneVision\mbox{-}7B}~\cite{llava}     
                                                        & 2024 & 16   & 97.7 & 82.0 & 70.1 & 23.6 & 28.5 & 46.5 & 58.1 \\
\model{Qwen2.5\mbox{-}VL\mbox{-}7B}~\cite{bai2025qwen25vltechnicalreport}  
                                                        & 2024 & 32   & 98.6 & 78.6 & 34.7 & 25.1 & 27.4 & 38.8 & 50.5 \\
\openrow  \model{VideoLLaMA2\mbox{-}7B}~\cite{cheng2024videollama2} & 2024 & 16   & 97.7 & 63.7 & 43.8 & 28.7 & 32.0 & 33.2 & 49.9 \\
\model{VideoLLaMA3\mbox{-}7B}~\cite{cheng2024videollama2} & 2025 & 180  & 99.1 & 83.5 & 58.1 & 26.1 & 29.8 & 46.9 & 57.3 \\
\openrow  \model{MiniCPM\mbox{-}V4.5}~\cite{hu2024minicpmunveilingpotentialsmall} 
                                                        & 2025 & 540  & 98.0 & 72.4 & 62.5 & 30.7 & 34.9 & 50.1 & 58.1 \\
\model{VideoR1\mbox{-}7B}~\cite{cheng2024videollama2}     & 2025 & 16   & 98.5 & 71.9 & 71.9 & 19.5 & 24.6 & 40.7 & 54.5 \\
\bottomrule
\end{tabular}}
\end{table*}

\begin{table*}[t]
\centering
\caption{Performance of different MLLMs on the \ourbench~benchmark.  The average improvement of Scene-RAG is $\sim$1.40\% in average.}
\label{tab:ourbench_rag}
\resizebox{0.95\textwidth}{!}{
\begin{tabular}{lc|c|c|cccccc|c|c}
\toprule
\rowcolor{headerpurple}
\begin{tabular}[c]{@{}c@{}}\textbf{Method}\end{tabular} &
\begin{tabular}[c]{@{}c@{}}\textbf{RAG}\\\textbf{Type}\end{tabular} &
\makecell{\textbf{LLM}\\\textbf{Params}} &
\makecell{\textbf{Frame}\\\textbf{Count}} &
\makecell{\textbf{Title}\\\textbf{Pred.}} &
\makecell{\textbf{Comm.}\\\textbf{Pred.}} &
\makecell{\textbf{Clip}\\\textbf{QA}} &
\makecell{\textbf{Scene}\\\textbf{QA}} &
\makecell{\textbf{SceneQA}\\\textbf{-Audio}} &
\makecell{\textbf{I-VQA}} &
\makecell{\textbf{Avg.}} &
\makecell{\textbf{Gain}} \\

\midrule
LongVA \cite{longva} & - & 7B & 128 & 96.8 & 71.1 & 54.3 & 27.7 & 29.5 & 40.1  & 53.3 & -  \\
LongVA \cite{longva} & + Video-RAG~\cite{luo2024video} & 7B & 128 
& 97.7 & 70.4 & 53.7 & 28.3 & 30.2 & 39.2 & 53.4 & +0.1 \\
LongVA \cite{longva} & + Scene-RAG & 7B & 128 
& 97.7 & 71.5 & 54.6 & 28.9 & 32.4 & 41.4 & 54.4 & +1.1 \\
\midrule
\openrow Long-LLaVA \cite{longllava} & - & 7B & 64 & 96.2 & 79.7 & 57.1 & 26.1 & 25.4 & 33.0 & 52.9 & - \\

\openrow Long-LLaVA \cite{longllava} & + Video-RAG~\cite{luo2024video} & 7B & 64 & 96.2  & 80.8 & 56.5 & 24.3 & 24.3 & 35.4 & 52.9 & +0.0 \\
\openrow Long-LLaVA \cite{longllava} & + Scene-RAG & 7B & 64 & 96.2 & 81.8 & 57.6 & 26.8 & 28.0 & 37.0 & 54.6 & +1.7 \\
\midrule
{LLaVA\mbox{-}OneVision\mbox{-}7B}~\cite{llava} & - & 7B & 16 & 97.7  & 82.0 & 70.1 & 23.6 & 28.5 & 46.5 & 58.1 & - \\
{LLaVA\mbox{-}OneVision\mbox{-}7B}~\cite{llava} & + Video-RAG~\cite{luo2024video} & 7B & 16 
& 98.3 & 84.4 & 69.1 & 25.1 & 29.3 & 43.7 & 58.3 & +0.2 \\
{LLaVA\mbox{-}OneVision\mbox{-}7B}~\cite{llava} & + Scene-RAG & 7B & 16 
& 98.3 & 85.4 & 71.9 & 25.6 & 31.2 & 44.1 & 59.4 & +1.3 \\


\bottomrule
\end{tabular}
}
\end{table*}

\section{Experiments}

In this section, we first present a comprehensive benchmark of 19 state-of-the-art MLLMs on \ourdataset, providing a clear snapshot of the current landscape of multimodal long-context reasoning (Sec.~\ref{exp:benchmark_anlysis}). 
A consistent pattern quickly emerges: as the context stretches, performance falls, the longer the query, the steeper the decline. 
This observation motivates the second part of our study, where we delve into \ourrag\ (Sec.~\ref{exp:ourrag_anlysis}), a mechanism designed to probe and alleviate the memory-loss phenomenon inherent in long-context processing. As illustrated in Figure~\ref{fig:teaser}, \ourrag\ offers a promising pathway for mitigating these challenges.


For each model, we set input frame length and video resolution to the officially recommended optimal configuration. For \ourrag, the video retriever is derived from InternVideo2~\cite{wang2024internvideo2} , initialized with the weight after its stage 2 training. The number of input frames is 4. For Scene Tiling, we set the TV-L1 regularization coefficient to $\lambda=1.5$, the minimum segment length to $L_{\min}(s)=3$, and $topk=10$. The retrieved frames are mixed with the model's uniformly sampled frames at a ratio of 0.5. We perform all experiments on a single H800 80G GPU. 

\begin{table*}[t]
\centering %
\caption{Performance of different MLLMs on the Video-MME~\cite{fu2025video}.}
\label{tab:videomme_results}
\resizebox{1.0\textwidth}{!}{
\begin{tabular}{lc|c|c|cccc|c|c}
\toprule
\rowcolor{headerpurple}
\textbf{Method} & \textbf{RAG Type} & \textbf{LLM Params} & \textbf{Frame} &
\textbf{Short} & \textbf{Medium} & \textbf{Long} & \textbf{Overall} & \textbf{Avg.} & \textbf{Gain} \\
\midrule
LongVA \cite{longva} & - & 7B & 128 & 60.9 & 49.3 & 44.0 & 51.4 & 51.4 & - \\
LongVA \cite{longva} & + Video-RAG~\cite{luo2024video} & 7B & 128 & 66.1 & 60.4 & 59.4 & 62.0 & 62.0 & +10.6 \\
LongVA \cite{longva} & + Scene-RAG & 7B & 128 & 66.9 & 60.6 & 59.7 & 63.4 & 62.4 & +11.0 \\
\midrule
\openrow Long-LLaVA \cite{longllava} & - & 7B & 64 & 60.3 & 51.4 & 44.1 & 52.0 & 52.0 & - \\
\openrow Long-LLaVA \cite{longllava} & + Video-RAG~\cite{luo2024video} & 7B & 64 & 66.4 & 60.2 & 59.8 & 62.1 & 62.1 & +10.1 \\
\openrow Long-LLaVA \cite{longllava} & + Scene-RAG & 7B & 64 & 68.0 & 62.1 & 61.4 & 64.8 & 63.8 & +11.8 \\
\midrule
{LLaVA\mbox{-}OneVision\mbox{-}7B}~\cite{llava} & - & 7B & 16 & 49.4 & 43.0 &  36.7 &  43.0 &  43.0 & - \\
{LLaVA\mbox{-}OneVision\mbox{-}7B}~\cite{llava} & + Video-RAG~\cite{luo2024video} & 7B & 16 & 56.6 & 47.4 & 46.0 & 50.0 & 50.0 & +7.0 \\
{LLaVA\mbox{-}OneVision\mbox{-}7B}~\cite{llava}   & + Scene-RAG & 7B & 16  & 57.1 & 48.2 & 46.4 & 50.6 & 50.6 & +8.6 \\
\bottomrule
\end{tabular}
}
\end{table*}

\begin{table*}[t] 
\centering
\caption{\textbf{\ourrag{} ablation study on \ourdataset{} and MLVU~\cite{zhou2024mlvu}.} Each column indicates whether the corresponding modality or component is used. Results are reported  with an additional Gain column showing improvements over the no-RAG baseline.}
\label{tab:ablation_combined_mlvu_results}
\setlength{\tabcolsep}{8pt}
\renewcommand\arraystretch{1.1}

\begin{subtable}[t]{0.46\textwidth}
\centering
\resizebox{\textwidth}{!}{
\begin{tabular}{l|ccc|c|c}
\toprule
\rowcolor{headerpurple}
\textbf{Model} & \textbf{Visual} & \textbf{Audio} & \textbf{Tiling} & \textbf{SceneBench} & \textbf{Gain} \\
\midrule
LongVA~\cite{longva} & \xmark & \xmark & \xmark & 53.3 & -- \\
LongVA~\cite{longva} & \cmark & \xmark & \xmark & 53.5 & +0.2 \\
LongVA~\cite{longva} & \cmark & \cmark & \xmark & 54.0 & +0.7 \\
LongVA~\cite{longva} & \cmark & \cmark & \cmark & 54.4 & +1.1  \\
\bottomrule
\end{tabular}}
\caption{Results on SceneBench.}
\end{subtable}
\hfill
\begin{subtable}[t]{0.48\textwidth}
\centering
\resizebox{\textwidth}{!}{
\begin{tabular}{l|ccc|c|c}
\toprule
\rowcolor{headerpurple}
\textbf{Model} & \textbf{Visual} & \textbf{Audio} & \textbf{Tiling} & \textbf{MLVU~\cite{zhou2024mlvu}} & \textbf{Gain} \\
\midrule
LLaVA-Video-7B~\cite{llavavideo} & \xmark & \xmark & \xmark & 70.8 & -- \\
LLaVA-Video-7B~\cite{llavavideo} & \cmark & \xmark & \xmark & 71.2 & +0.4 \\
LLaVA-Video-7B~\cite{llavavideo} & \cmark & \cmark & \xmark & 73.4 & +2.6 \\
LLaVA-Video-7B~\cite{llavavideo} & \cmark & \cmark & \cmark & 74.1 & +3.3 \\
\bottomrule
\end{tabular}}
\caption{Results on MLVU~\cite{zhou2024mlvu}.}
\end{subtable}

\vspace{-0.5em}
\end{table*}

\subsection{\ourdataset~Benchmark}
\label{exp:benchmark_anlysis}
Table~\ref{tab:mllm_2024plus} and Table~\ref{tab:ourbench_rag} summarize the performance of 16 MLLMs across six tasks on \ourdataset.

We begin by examining the holistic video understanding tasks, namely title prediction (TP) and comment prediction (CP). Most models perform well on TP: they successfully capture the overall semantic intent of the video, and the majority reach accuracy above 90. In contrast, CP is substantially more challenging. This task requires predicting user comments that frequently show semantic divergence from the video content and therefore cannot be solved through simple pattern matching. As a consequence, the performance drops significantly for almost all models, with 15 out of 16 showing a clear decline. The magnitude of this decline ranges from about -15 for LLAVA-OneVision~\cite{llava}  to about -34 for Qen2.5-VL~\cite{qwen2.5-VL}.




Next, we examine performance on the clip-level task (ClipQA) and the scene-level tasks (SceneQA and SceneQA-Audio), where the latter require longer temporal reasoning and, in the case of SceneQA-Audio, additional integration of audio cues. As shown in Table~\ref{tab:mllm_2024plus}, a clear performance gap emerges between clip-level and scene-level understanding. Across the open-source MLLMs, the average scores are 51.73 for ClipQA, 26.11 for SceneQA, and 29.08 for SceneQA-Audio. On average, SceneQA drops by 25.62 points relative to ClipQA, corresponding to a 49.5\% decrease, while SceneQA-Audio remains 22.65 points below ClipQA but is 2.97 points higher than SceneQA. These results indicate that long-range scene-level reasoning is substantially more challenging than clip-level understanding. At the same time, the slight improvement of SceneQA-Audio over SceneQA suggests that audio cues can provide useful complementary evidence for part of the benchmark.

Finally, we examine I-VQA, which requires bidirectional reasoning across earlier and later plot points and thus depends heavily on long-range context. With four answer options, random guessing yields 25\% accuracy. The average I-VQA performance across the open-source MLLMs is 35.79. Notably, several models perform close to or even below the random-guessing baseline, suggesting reasoning is still challenging for current long-video MLLMs. 

The average performance across all models is 52.3 (Table~\ref{tab:mllm_2024plus}). Notably, several models perform near or even below the random-guessing baseline, indicating that capturing the required cross-scene dependencies remains highly challenging. In contrast, stronger proprietary models, especially Gemini 2.5 Pro, achieve substantially better results on this task, with Gemini 2.5 Pro reaching 80.3. Overall, these results suggest that robust bidirectional long-context reasoning in long videos is still far from solved, particularly for open-source models.


\subsection{\ourrag~and Memory Loss}
\label{exp:ourrag_anlysis}

To study long-term memory loss in MLLMs and evaluate how retrieval can mitigate it, we revisit the results in Table~\ref{tab:ourbench_rag} using three representative baselines~\cite{longva, longllava, llava} equipped with Video-RAG~\cite{luo2024video} and our proposed \ourrag.

Across all three models, Scene-RAG improves over the non-RAG baseline by an average of +0.50 on Title Prediction, +1.97 on Comment Prediction, +0.87 on ClipQA, +1.30 on SceneQA, +2.73 on SceneQA-Audio, and +0.97 on I-VQA. Relative to Video-RAG, Scene-RAG provides additional mean gains of +0.00, +1.03, +1.60, +1.20, +2.60, and +1.40, respectively. Notably, the most pronounced improvement is observed on SceneQA-Audio, while gains on SceneQA and ClipQA are also substantial, suggesting that scene-structured retrieval is especially beneficial for tasks requiring long-range evidence aggregation.

As shown in Figure~\ref{fig:teaser}, we analyze how accuracy varies with increasing temporal distance between questions and supporting evidence. We compare the baseline, Video-RAG, and Scene-RAG. Scene-RAG maintains accuracy at medium and long distances, better mitigating long-context forgetting than frame-level retrieval.




\subsection{\ourrag~Ablation}
\label{exp:ourrag_ablation}

We further validate \ourrag{} by applying the same RAG comparison setting on Video-MME~\cite{fu2025video}. As shown in Table~\ref{tab:videomme_results}, Video-RAG improves the average score by 9.23 points over the non-RAG baseline across the three MLLMs, while \ourrag{} achieves a larger average gain of 10.47 points. Moreover, \ourrag{} consistently outperforms Video-RAG on all three models, further demonstrating its robustness on long-video understanding.

Next, we analyze the contribution of each component in \ourrag{} through an ablation study. Starting from the baseline model, we gradually enable (1) visual retrieval only, (2) visual + audio retrieval, and (3) the full model with visual retrieval, audio retrieval, and SceneTiling. This ablation is conducted on both \ourdataset{} and MLVU~\cite{zhou2024mlvu}, with results shown in Table~\ref{tab:ablation_combined_mlvu_results}. 


\section{Conclusion}

In this work, we present SceneBench, a new scene-centric benchmark designed to reveal key challenges in long-term video understanding. Evaluations on SceneBench expose a substantial performance drop when models are required to reason over long contexts (the longer, the worse), highlighting the limitations of current MLLMs in long-term memory, cross-scene evidence aggregation, and causal inference.


To mitigate these issues, we further propose Scene-RAG, a scene-structured retrieval framework that constructs multimodal scene memory to support long-context reasoning. Experimental results across multiple benchmarks demonstrate that Scene-RAG not only achieves robust improvements over existing retrieval approaches, but also effectively mitigates long-context forgetting in MLLMs. 

We believe that SceneBench and Scene-RAG will serve as valuable resources for advancing research toward MLLMs with stronger scene understanding and more reliable long-term video comprehension.




{
    \small
    \bibliographystyle{ieeenat_fullname}
    \bibliography{main}

\begin{thebibliography}{72}
\providecommand{\natexlab}[1]{#1}
\providecommand{\url}[1]{\texttt{#1}}
\expandafter\ifx\csname urlstyle\endcsname\relax
  \providecommand{\doi}[1]{doi: #1}\else
  \providecommand{\doi}{doi: \begingroup \urlstyle{rm}\Url}\fi

\bibitem[Bai et~al.(2025{\natexlab{a}})Bai, Chen, Liu, Wang, Ge, Song, Dang, Wang, Wang, Tang, Zhong, Zhu, Yang, Li, Wan, Wang, Ding, Fu, Xu, Ye, Zhang, Xie, Cheng, Zhang, Yang, Xu, and Lin]{bai2025qwen25vltechnicalreport}
Shuai Bai, Keqin Chen, Xuejing Liu, Jialin Wang, Wenbin Ge, Sibo Song, Kai Dang, Peng Wang, Shijie Wang, Jun Tang, Humen Zhong, Yuanzhi Zhu, Mingkun Yang, Zhaohai Li, Jianqiang Wan, Pengfei Wang, Wei Ding, Zheren Fu, Yiheng Xu, Jiabo Ye, Xi Zhang, Tianbao Xie, Zesen Cheng, Hang Zhang, Zhibo Yang, Haiyang Xu, and Junyang Lin.
\newblock Qwen2.5-vl technical report, 2025{\natexlab{a}}.

\bibitem[Bai et~al.(2025{\natexlab{b}})Bai, Chen, Liu, Wang, Ge, Song, Dang, Wang, Wang, Tang, et~al.]{qwen2.5-VL}
Shuai Bai, Keqin Chen, Xuejing Liu, Jialin Wang, Wenbin Ge, Sibo Song, Kai Dang, Peng Wang, Shijie Wang, Jun Tang, et~al.
\newblock Qwen2. 5-vl technical report.
\newblock \emph{arXiv preprint arXiv:2502.13923}, 2025{\natexlab{b}}.

\bibitem[Chandrasegaran et~al.(2024)Chandrasegaran, Gupta, Hadzic, Kota, He, Eyzaguirre, Durante, Li, Wu, and Fei-Fei]{chandrasegaran2024hourvideo}
Keshigeyan Chandrasegaran, Agrim Gupta, Lea~M Hadzic, Taran Kota, Jimming He, Crist{\'o}bal Eyzaguirre, Zane Durante, Manling Li, Jiajun Wu, and Li Fei-Fei.
\newblock Hourvideo: 1-hour video-language understanding.
\newblock \emph{Advances in Neural Information Processing Systems}, 37:\penalty0 53168--53197, 2024.

\bibitem[Chen et~al.(2024{\natexlab{a}})Chen, Hou, Qu, Testini, Hong, and Jiao]{chen2024360}
Hao Chen, Yuqi Hou, Chenyuan Qu, Irene Testini, Xiaohan Hong, and Jianbo Jiao.
\newblock 360+x: A panoptic multi-modal scene understanding dataset.
\newblock In \emph{Proceedings of the IEEE/CVF Conference on Computer Vision and Pattern Recognition}, pages 19373--19382, 2024{\natexlab{a}}.

\bibitem[Chen et~al.(2024{\natexlab{b}})Chen, Wang, Cao, Liu, Gao, Cui, Zhu, Ye, Tian, Liu, et~al.]{chen2024expanding}
Zhe Chen, Weiyun Wang, Yue Cao, Yangzhou Liu, Zhangwei Gao, Erfei Cui, Jinguo Zhu, Shenglong Ye, Hao Tian, Zhaoyang Liu, et~al.
\newblock Expanding performance boundaries of open-source multimodal models with model, data, and test-time scaling.
\newblock \emph{arXiv preprint arXiv:2412.05271}, 2024{\natexlab{b}}.

\bibitem[Chen et~al.(2024{\natexlab{c}})Chen, Wu, Wang, Su, Chen, Xing, Zhong, Zhang, Zhu, Lu, et~al.]{chen2024internvl}
Zhe Chen, Jiannan Wu, Wenhai Wang, Weijie Su, Guo Chen, Sen Xing, Muyan Zhong, Qinglong Zhang, Xizhou Zhu, Lewei Lu, et~al.
\newblock Internvl: Scaling up vision foundation models and aligning for generic visual-linguistic tasks.
\newblock In \emph{Proceedings of the IEEE/CVF conference on computer vision and pattern recognition}, pages 24185--24198, 2024{\natexlab{c}}.

\bibitem[Cheng et~al.(2024)Cheng, Leng, Zhang, Xin, Li, Chen, Zhu, Zhang, Luo, Zhao, et~al.]{cheng2024videollama2}
Zesen Cheng, Sicong Leng, Hang Zhang, Yifei Xin, Xin Li, Guanzheng Chen, Yongxin Zhu, Wenqi Zhang, Ziyang Luo, Deli Zhao, et~al.
\newblock Videollama 2: Advancing spatial-temporal modeling and audio understanding in video-llms.
\newblock \emph{arXiv preprint arXiv:2406.07476}, 2024.

\bibitem[Chu et~al.(2024)Chu, Xu, Yang, Wei, Wei, Guo, Leng, Lv, He, Lin, et~al.]{chu2024qwen2}
Yunfei Chu, Jin Xu, Qian Yang, Haojie Wei, Xipin Wei, Zhifang Guo, Yichong Leng, Yuanjun Lv, Jinzheng He, Junyang Lin, et~al.
\newblock Qwen2-audio technical report.
\newblock \emph{arXiv preprint arXiv:2407.10759}, 2024.

\bibitem[Comanici et~al.(2025)Comanici, Bieber, and et~al.]{comanici2025gemini25pushingfrontier}
Gheorghe Comanici, Eric Bieber, and Mike~Schaekermann et al.
\newblock Gemini 2.5: Pushing the frontier with advanced reasoning, multimodality, long context, and next generation agentic capabilities, 2025.

\bibitem[Fei et~al.(2024)Fei, Li, Deng, Wang, Liu, and Wang]{videoccam}
Jiajun Fei, Dian Li, Zhidong Deng, Zekun Wang, Gang Liu, and Hui Wang.
\newblock Video-ccam: Enhancing video-language understanding with causal cross-attention masks for short and long videos.
\newblock \emph{arXiv preprint arXiv:2408.14023}, 2024.

\bibitem[Fu et~al.(2023)Fu, Chen, Shen, Qin, Zhang, Lin, Yang, Zheng, Li, Sun, Wu, and Ji]{fu2023mme}
Chaoyou Fu, Peixian Chen, Yunhang Shen, Yulei Qin, Mengdan Zhang, Xu Lin, Jinrui Yang, Xiawu Zheng, Ke Li, Xing Sun, Yunsheng Wu, and Rongrong Ji.
\newblock Mme: A comprehensive evaluation benchmark for multimodal large language models.
\newblock \emph{arXiv preprint arXiv:2306.13394}, 2023.

\bibitem[Fu et~al.(2024{\natexlab{a}})Fu, Dai, Luo, Li, Ren, Zhang, Wang, Zhou, Shen, Zhang, Chen, Li, Lin, Zhao, Li, Xu, Zheng, Chen, Ji, and Sun]{fu2024videommefirstevercomprehensiveevaluation}
Chaoyou Fu, Yuhan Dai, Yongdong Luo, Lei Li, Shuhuai Ren, Renrui Zhang, Zihan Wang, Chenyu Zhou, Yunhang Shen, Mengdan Zhang, Peixian Chen, Yanwei Li, Shaohui Lin, Sirui Zhao, Ke Li, Tong Xu, Xiawu Zheng, Enhong Chen, Rongrong Ji, and Xing Sun.
\newblock Video-mme: The first-ever comprehensive evaluation benchmark of multi-modal llms in video analysis, 2024{\natexlab{a}}.

\bibitem[Fu et~al.(2024{\natexlab{b}})Fu, Dai, Luo, Li, Ren, Zhang, Wang, Zhou, Shen, Zhang, et~al.]{videomme}
Chaoyou Fu, Yuhan Dai, Yondong Luo, Lei Li, Shuhuai Ren, Renrui Zhang, Zihan Wang, Chenyu Zhou, Yunhang Shen, Mengdan Zhang, et~al.
\newblock Video-mme: The first-ever comprehensive evaluation benchmark of multi-modal llms in video analysis.
\newblock \emph{arXiv preprint arXiv:2405.21075}, 2024{\natexlab{b}}.

\bibitem[Fu et~al.(2025)Fu, Dai, Luo, Li, Ren, Zhang, Wang, Zhou, Shen, Zhang, et~al.]{fu2025video}
Chaoyou Fu, Yuhan Dai, Yongdong Luo, Lei Li, Shuhuai Ren, Renrui Zhang, Zihan Wang, Chenyu Zhou, Yunhang Shen, Mengdan Zhang, et~al.
\newblock Video-mme: The first-ever comprehensive evaluation benchmark of multi-modal llms in video analysis.
\newblock In \emph{Proceedings of the Computer Vision and Pattern Recognition Conference}, pages 24108--24118, 2025.

\bibitem[Guo et~al.(2024)Guo, Zheng, Bai, Li, Wang, Zhu, Li, Neubig, Chen, and Yue]{guo2024mammoth}
Jarvis Guo, Tuney Zheng, Yuelin Bai, Bo Li, Yubo Wang, King Zhu, Yizhi Li, Graham Neubig, Wenhu Chen, and Xiang Yue.
\newblock Mammoth-vl: Eliciting multimodal reasoning with instruction tuning at scale.
\newblock \emph{arXiv preprint arXiv:2412.05237}, 2024.

\bibitem[Hasson et~al.(2008)Hasson, Landesman, Knappmeyer, Vallines, Rubin, and Heeger]{hasson2008neurocinematics}
Uri Hasson, Ohad Landesman, Barbara Knappmeyer, Ignacio Vallines, Nava Rubin, and David~J Heeger.
\newblock Neurocinematics: The neuroscience of film.
\newblock \emph{Projections}, 2\penalty0 (1):\penalty0 1--26, 2008.

\bibitem[He et~al.(2024)He, Li, Jang, Jia, Cao, Shah, Shrivastava, and Lim]{malmm2024}
Bo He, Hengduo Li, Young~Kyun Jang, Menglin Jia, Xuefei Cao, Ashish Shah, Abhinav Shrivastava, and Ser-Nam Lim.
\newblock Ma-lmm: Memory-augmented large multimodal model for long-term video understanding.
\newblock \emph{arXiv preprint arXiv:2404.05726}, 2024.

\bibitem[Hu et~al.(2024)Hu, Tu, Han, He, Cui, Long, Zheng, Fang, Huang, Zhao, Zhang, Thai, Zhang, Wang, Yao, Zhao, Zhou, Cai, Zhai, Ding, Jia, Zeng, Li, Liu, and Sun]{hu2024minicpmunveilingpotentialsmall}
Shengding Hu, Yuge Tu, Xu Han, Chaoqun He, Ganqu Cui, Xiang Long, Zhi Zheng, Yewei Fang, Yuxiang Huang, Weilin Zhao, Xinrong Zhang, Zheng~Leng Thai, Kaihuo Zhang, Chongyi Wang, Yuan Yao, Chenyang Zhao, Jie Zhou, Jie Cai, Zhongwu Zhai, Ning Ding, Chao Jia, Guoyang Zeng, Dahai Li, Zhiyuan Liu, and Maosong Sun.
\newblock Minicpm: Unveiling the potential of small language models with scalable training strategies, 2024.

\bibitem[Jeong et~al.(2025)Jeong, Kim, Baek, and Hwang]{jeong2025videoragretrievalaugmentedgenerationvideo}
Soyeong Jeong, Kangsan Kim, Jinheon Baek, and Sung~Ju Hwang.
\newblock Videorag: Retrieval-augmented generation over video corpus, 2025.

\bibitem[Khattak et~al.(2024)Khattak, Naeem, Hassan, Naseer, Tombari, Khan, and Khan]{cvrr-2024}
Muhammad~Uzair Khattak, Muhammad~Ferjad Naeem, Jameel Hassan, Muzammal Naseer, Federico Tombari, Fahad~Shahbaz Khan, and Salman Khan.
\newblock Complex video reasoning and robustness evaluation suite for video-lmms.
\newblock \emph{arXiv preprint arXiv:2405.03690}, 2024.

\bibitem[Li et~al.(2023{\natexlab{a}})Li, Wang, Wang, Ge, Ge, and Shan]{seedbench2023}
Bohao Li, Rui Wang, Guangzhi Wang, Yuying Ge, Yixiao Ge, and Ying Shan.
\newblock Seed-bench: Benchmarking multimodal llms with generative comprehension.
\newblock \emph{arXiv preprint arXiv:2307.16125}, 2023{\natexlab{a}}.

\bibitem[Li et~al.(2023{\natexlab{b}})Li, Wang, He, Li, Wang, Liu, Wang, Xu, Chen, Luo, et~al.]{li2023mvbench}
Kunchang Li, Yali Wang, Yinan He, Yizhuo Li, Yi Wang, Yi Liu, Zun Wang, Jilan Xu, Guo Chen, Ping Luo, et~al.
\newblock Mvbench: A comprehensive multi-modal video understanding benchmark.
\newblock \emph{ArXiv preprint}, 2023{\natexlab{b}}.

\bibitem[Li et~al.(2023{\natexlab{c}})Li, Wang, He, Li, Wang, Liu, Wang, Xu, Chen, Luo, et~al.]{mvbench2023}
Kunchang Li, Yali Wang, Yinan He, Yizhuo Li, Yi Wang, Yi Liu, Zun Wang, Jilan Xu, Guo Chen, Ping Luo, et~al.
\newblock Mvbench: A comprehensive multi-modal video understanding benchmark.
\newblock \emph{arXiv preprint arXiv:2311.17005}, 2023{\natexlab{c}}.

\bibitem[Li et~al.(2023{\natexlab{d}})Li, Wang, and Jia]{llama-vid2023}
Yanwei Li, Chengyao Wang, and Jiaya Jia.
\newblock Llama-vid: An image is worth 2 tokens in large language models.
\newblock \emph{arXiv preprint arXiv:2311.17043}, 2023{\natexlab{d}}.

\bibitem[Li et~al.(2024)Li, Chen, Hu, Wang, Shi, and Zhang]{li2024videovista}
Yunxin Li, Xinyu Chen, Baotian Hu, Longyue Wang, Haoyuan Shi, and Min Zhang.
\newblock Videovista: A versatile benchmark for video understanding and reasoning.
\newblock \emph{arXiv preprint arXiv:2406.11303}, 2024.

\bibitem[Lin et~al.(2023)Lin, Zhu, Ye, Ning, Jin, and Yuan]{videollava2023}
Bin Lin, Bin Zhu, Yang Ye, Munan Ning, Peng Jin, and Li Yuan.
\newblock Video-llava: Learning united visual representation by alignment before projection.
\newblock \emph{arXiv preprint arXiv:2311.10122}, 2023.

\bibitem[Liu et~al.(2023{\natexlab{a}})Liu, Li, Wu, and Lee]{liu2023visual}
Haotian Liu, Chunyuan Li, Qingyang Wu, and Yong~Jae Lee.
\newblock Visual instruction tuning.
\newblock \emph{Advances in neural information processing systems}, 36:\penalty0 34892--34916, 2023{\natexlab{a}}.

\bibitem[Liu et~al.(2023{\natexlab{b}})Liu, Li, Wu, and Lee]{llava}
Haotian Liu, Chunyuan Li, Qingyang Wu, and Yong~Jae Lee.
\newblock Visual instruction tuning, 2023{\natexlab{b}}.

\bibitem[Liu et~al.(2024)Liu, Yan, Zaharia, and Abbeel]{liu2024world}
Hao Liu, Wilson Yan, Matei Zaharia, and Pieter Abbeel.
\newblock World model on million-length video and language with ringattention.
\newblock \emph{arXiv preprint arXiv:2402.08268}, 2024.

\bibitem[Liu et~al.(2023{\natexlab{c}})Liu, Duan, Zhang, Li, Zhang, Zhao, Yuan, Wang, He, Liu, et~al.]{liu2023mmbench}
Yuan Liu, Haodong Duan, Yuanhan Zhang, Bo Li, Songyang Zhang, Wangbo Zhao, Yike Yuan, Jiaqi Wang, Conghui He, Ziwei Liu, et~al.
\newblock Mmbench: Is your multi-modal model an all-around player?
\newblock \emph{arXiv preprint arXiv:2307.06281}, 2023{\natexlab{c}}.

\bibitem[Luo et~al.(2024{\natexlab{a}})Luo, Zheng, Yang, Li, Lin, Huang, Ji, Chao, Luo, and Ji]{luo2024video}
Yongdong Luo, Xiawu Zheng, Xiao Yang, Guilin Li, Haojia Lin, Jinfa Huang, Jiayi Ji, Fei Chao, Jiebo Luo, and Rongrong Ji.
\newblock Video-rag: Visually-aligned retrieval-augmented long video comprehension.
\newblock \emph{arXiv preprint arXiv:2411.13093}, 2024{\natexlab{a}}.

\bibitem[Luo et~al.(2024{\natexlab{b}})Luo, Zheng, Yang, Li, Lin, Huang, Ji, Chao, Luo, and Ji]{luo2024videoragvisuallyalignedretrievalaugmentedlong}
Yongdong Luo, Xiawu Zheng, Xiao Yang, Guilin Li, Haojia Lin, Jinfa Huang, Jiayi Ji, Fei Chao, Jiebo Luo, and Rongrong Ji.
\newblock Video-rag: Visually-aligned retrieval-augmented long video comprehension, 2024{\natexlab{b}}.

\bibitem[Ma et~al.(2025{\natexlab{a}})Ma, Yuan, Wang, Zang, Liu, He, Wei, Guo, Jiahui, Yang, et~al.]{ma2025scalelong}
David Ma, Huaqing Yuan, Xingjian Wang, Qianbo Zang, Tianci Liu, Xinyang He, Yanbin Wei, Jiawei Guo, Ni Jiahui, Zhenzhu Yang, et~al.
\newblock Scalelong: A multi-timescale benchmark for long video understanding.
\newblock \emph{arXiv preprint arXiv:2505.23922}, 2025{\natexlab{a}}.

\bibitem[Ma et~al.(2025{\natexlab{b}})Ma, Zhang, Ren, Guo, Yao, Wei, Yang, Peng, Feng, Ma, et~al.]{ma2025iv}
David Ma, Yuanxing Zhang, Jincheng Ren, Jarvis Guo, Yifan Yao, Zhenlin Wei, Zhenzhu Yang, Zhongyuan Peng, Boyu Feng, Jun Ma, et~al.
\newblock Iv-bench: A benchmark for image-grounded video perception and reasoning in multimodal llms.
\newblock \emph{arXiv preprint arXiv:2504.15415}, 2025{\natexlab{b}}.

\bibitem[Maaz et~al.(2023)Maaz, Rasheed, Khan, and Khan]{videochatgpt2023}
Muhammad Maaz, Hanoona Rasheed, Salman Khan, and Fahad~Shahbaz Khan.
\newblock Video-chatgpt: Towards detailed video understanding via large vision and language models.
\newblock \emph{arXiv preprint arXiv:2306.05424}, 2023.

\bibitem[Mariola et~al.(2022)Mariola, Fountas, Barnett, and Roseboom]{mariola2022event}
Alberto Mariola, Zafeirios Fountas, Lionel Barnett, and Warrick Roseboom.
\newblock Event segmentation in continuous, naturalistic videos from model-based, data-driven, and human perspectives.
\newblock 2022.

\bibitem[Miech et~al.(2019)Miech, Zhukov, Alayrac, Tapaswi, Laptev, and Sivic]{howto100m2019}
Antoine Miech, Dimitri Zhukov, Jean-Baptiste Alayrac, Makarand Tapaswi, Ivan Laptev, and Josef Sivic.
\newblock Howto100m: Learning a text-video embedding by watching hundred million narrated video clips.
\newblock In \emph{Proceedings of the IEEE/CVF international conference on computer vision}, pages 2630--2640, 2019.

\bibitem[Ren et~al.(2023)Ren, Yao, Li, Sun, and Hou]{timechat}
Shuhuai Ren, Linli Yao, Shicheng Li, Xu Sun, and Lu Hou.
\newblock Timechat: A time-sensitive multimodal large language model for long video understanding.
\newblock \emph{ArXiv preprint}, 2023.

\bibitem[Shu et~al.(2024)Shu, Zhang, Liu, Qin, Zhou, Huang, and Zhao]{shu2024videoxl}
Yan Shu, Peitian Zhang, Zheng Liu, Minghao Qin, Junjie Zhou, Tiejun Huang, and Bo Zhao.
\newblock Video-xl: Extra-long vision language model for hour-scale video understanding.
\newblock \emph{arXiv preprint arXiv:2409.14485}, 2024.

\bibitem[Smith(2012)]{smith2012attentional}
TJ Smith.
\newblock The attentional theory of cinematic continuity. projections, 6 (1), 1-27.
\newblock \emph{Berghahn Journals}, 2012.

\bibitem[Song et~al.(2023{\natexlab{a}})Song, Chai, Wang, Zhang, Zhou, Wu, Guo, Ye, Lu, Hwang, et~al.]{moviechat}
Enxin Song, Wenhao Chai, Guanhong Wang, Yucheng Zhang, Haoyang Zhou, Feiyang Wu, Xun Guo, Tian Ye, Yan Lu, Jenq-Neng Hwang, et~al.
\newblock Moviechat: From dense token to sparse memory for long video understanding.
\newblock \emph{arXiv preprint arXiv:2307.16449}, 2023{\natexlab{a}}.

\bibitem[Song et~al.(2023{\natexlab{b}})Song, Chai, Wang, Zhang, Zhou, Wu, Guo, Ye, Lu, Hwang, et~al.]{moviechat2023}
Enxin Song, Wenhao Chai, Guanhong Wang, Yucheng Zhang, Haoyang Zhou, Feiyang Wu, Xun Guo, Tian Ye, Yan Lu, Jenq-Neng Hwang, et~al.
\newblock Moviechat: From dense token to sparse memory for long video understanding.
\newblock \emph{arXiv preprint arXiv:2307.16449}, 2023{\natexlab{b}}.

\bibitem[Tan et~al.(2025)Tan, Luo, Ye, Liu, and Cai]{tan2025allvb}
Xichen Tan, Yuanjing Luo, Yunfan Ye, Fang Liu, and Zhiping Cai.
\newblock Allvb: All-in-one long video understanding benchmark.
\newblock In \emph{Proceedings of the AAAI Conference on Artificial Intelligence}, pages 7211--7219, 2025.

\bibitem[Team et~al.(2025)Team, Du, Yin, Xing, Qu, Wang, Chen, Zhang, Du, Wei, et~al.]{team2025kimi}
Kimi Team, Angang Du, Bohong Yin, Bowei Xing, Bowen Qu, Bowen Wang, Cheng Chen, Chenlin Zhang, Chenzhuang Du, Chu Wei, et~al.
\newblock Kimi-vl technical report.
\newblock \emph{arXiv preprint arXiv:2504.07491}, 2025.

\bibitem[Team et~al.(2026)Team, Bai, and et~al.]{kimiteam2026kimik25visualagentic}
Kimi Team, Tongtong Bai, and Yifan~Bai et al.
\newblock Kimi k2.5: Visual agentic intelligence, 2026.

\bibitem[Trabasso et~al.(1982)]{trabasso1982causal}
Tom Trabasso et~al.
\newblock Causal cohesion and story coherence.
\newblock 1982.

\bibitem[Wang et~al.(2025)Wang, He, Hong, Cheng, Zhang, Qi, Ding, Gu, Huang, Xu, et~al.]{wang2025lvbench}
Weihan Wang, Zehai He, Wenyi Hong, Yean Cheng, Xiaohan Zhang, Ji Qi, Ming Ding, Xiaotao Gu, Shiyu Huang, Bin Xu, et~al.
\newblock Lvbench: An extreme long video understanding benchmark.
\newblock In \emph{Proceedings of the IEEE/CVF International Conference on Computer Vision}, pages 22958--22967, 2025.

\bibitem[Wang et~al.(2024{\natexlab{a}})Wang, Zhang, Zohar, and Yeung-Levy]{videoagent2024}
Xiaohan Wang, Yuhui Zhang, Orr Zohar, and Serena Yeung-Levy.
\newblock Videoagent: Long-form video understanding with large language model as agent.
\newblock \emph{arXiv preprint arXiv:2403.10517}, 2024{\natexlab{a}}.

\bibitem[Wang et~al.(2024{\natexlab{b}})Wang, Li, Li, Yu, He, Chen, Pei, Zheng, Wang, Shi, et~al.]{wang2024internvideo2}
Yi Wang, Kunchang Li, Xinhao Li, Jiashuo Yu, Yinan He, Guo Chen, Baoqi Pei, Rongkun Zheng, Zun Wang, Yansong Shi, et~al.
\newblock Internvideo2: Scaling foundation models for multimodal video understanding.
\newblock In \emph{European Conference on Computer Vision}, pages 396--416. Springer, 2024{\natexlab{b}}.

\bibitem[Wu et~al.(2021)Wu, Yu, Chen, Tenenbaum, and Gan]{star2021}
Bo Wu, Shoubin Yu, Zhenfang Chen, Joshua~B Tenenbaum, and Chuang Gan.
\newblock Star: A benchmark for situated reasoning in real-world videos.
\newblock In \emph{Thirty-fifth conference on neural information processing systems datasets and benchmarks track (Round 2)}, 2021.

\bibitem[{Wu} et~al.(2024){Wu}, {Yu}, {Chen}, {Tenenbaum}, and {Gan}]{2024arXiv240509711W}
Bo {Wu}, Shoubin {Yu}, Zhenfang {Chen}, Joshua~B {Tenenbaum}, and Chuang {Gan}.
\newblock {STAR: A Benchmark for Situated Reasoning in Real-World Videos}.
\newblock \emph{arXiv e-prints}, art. arXiv:2405.09711, 2024.

\bibitem[Wu et~al.(2024)Wu, Li, Chen, and Li]{wu2024longvideobench}
Haoning Wu, Dongxu Li, Bei Chen, and Junnan Li.
\newblock Longvideobench: A benchmark for long-context interleaved video-language understanding.
\newblock \emph{Advances in Neural Information Processing Systems}, 37:\penalty0 28828--28857, 2024.

\bibitem[Xu et~al.(2016)Xu, Mei, Yao, and Rui]{msrvtt2016}
Jun Xu, Tao Mei, Ting Yao, and Yong Rui.
\newblock Msr-vtt: A large video description dataset for bridging video and language.
\newblock In \emph{Proceedings of the IEEE conference on computer vision and pattern recognition}, pages 5288--5296, 2016.

\bibitem[Xu et~al.(2023)Xu, Lan, Xie, Chen, and Lu]{R-VLM-2023}
Jiaqi Xu, Cuiling Lan, Wenxuan Xie, Xuejin Chen, and Yan Lu.
\newblock Retrieval-based video language model for efficient long video question answering.
\newblock \emph{arXiv preprint arXiv:2312.04931}, 2023.

\bibitem[Yang et~al.(2025)Yang, Li, Yang, Zhang, Hui, Zheng, Yu, Gao, Huang, Lv, et~al.]{yang2025qwen3}
An Yang, Anfeng Li, Baosong Yang, Beichen Zhang, Binyuan Hui, Bo Zheng, Bowen Yu, Chang Gao, Chengen Huang, Chenxu Lv, et~al.
\newblock Qwen3 technical report.
\newblock \emph{arXiv preprint arXiv:2505.09388}, 2025.

\bibitem[Ye et~al.(2024)Ye, Xu, Liu, Hu, Yan, Qian, Zhang, Huang, and Zhou]{ye2024mplug}
Jiabo Ye, Haiyang Xu, Haowei Liu, Anwen Hu, Ming Yan, Qi Qian, Ji Zhang, Fei Huang, and Jingren Zhou.
\newblock mplug-owl3: Towards long image-sequence understanding in multi-modal large language models.
\newblock \emph{arXiv preprint arXiv:2408.04840}, 2024.

\bibitem[Ye et~al.(2023)Ye, Xu, Xu, Ye, Yan, Zhou, Wang, Hu, Shi, Shi, et~al.]{mplug-owl-2023}
Qinghao Ye, Haiyang Xu, Guohai Xu, Jiabo Ye, Ming Yan, Yiyang Zhou, Junyang Wang, Anwen Hu, Pengcheng Shi, Yaya Shi, et~al.
\newblock mplug-owl: Modularization empowers large language models with multimodality.
\newblock \emph{arXiv preprint arXiv:2304.14178}, 2023.

\bibitem[{Yin Song and Chen Wu and Eden Duthie}(2024)]{longllava}
{Yin Song and Chen Wu and Eden Duthie}.
\newblock {aws-prototyping/long-llava-qwen2-7b}, 2024.

\bibitem[Yu et~al.(2019)Yu, Xu, Yu, Yu, Zhao, Zhuang, and Tao]{activitynetqa2019}
Zhou Yu, Dejing Xu, Jun Yu, Ting Yu, Zhou Zhao, Yueting Zhuang, and Dacheng Tao.
\newblock Activitynet-qa: A dataset for understanding complex web videos via question answering.
\newblock In \emph{Proceedings of the AAAI Conference on Artificial Intelligence}, pages 9127--9134, 2019.

\bibitem[Yuan et~al.(2025)Yuan, Liu, Qin, Qian, Shu, Dou, Wen, and Sebe]{yuan2025memoryenhancedretrievalaugmentationlong}
Huaying Yuan, Zheng Liu, Minghao Qin, Hongjin Qian, Yan Shu, Zhicheng Dou, Ji-Rong Wen, and Nicu Sebe.
\newblock Memory-enhanced retrieval augmentation for long video understanding, 2025.

\bibitem[Yue et~al.(2024)Yue, Ni, Zhang, Zheng, Liu, Zhang, Stevens, Jiang, Ren, Sun, Wei, Yu, Yuan, Sun, Yin, Zheng, Yang, Liu, Huang, Sun, Su, and Chen]{yue2023mmmu}
Xiang Yue, Yuansheng Ni, Kai Zhang, Tianyu Zheng, Ruoqi Liu, Ge Zhang, Samuel Stevens, Dongfu Jiang, Weiming Ren, Yuxuan Sun, Cong Wei, Botao Yu, Ruibin Yuan, Renliang Sun, Ming Yin, Boyuan Zheng, Zhenzhu Yang, Yibo Liu, Wenhao Huang, Huan Sun, Yu Su, and Wenhu Chen.
\newblock Mmmu: A massive multi-discipline multimodal understanding and reasoning benchmark for expert agi.
\newblock In \emph{Proceedings of CVPR}, 2024.

\bibitem[Zacks and Tversky(2001)]{zacks2001event}
Jeffrey~M Zacks and Barbara Tversky.
\newblock Event structure in perception and conception.
\newblock \emph{Psychological bulletin}, 127\penalty0 (1):\penalty0 3, 2001.

\bibitem[Zacks et~al.(2007)Zacks, Speer, Swallow, Braver, and Reynolds]{zacks2007event}
Jeffrey~M Zacks, Nicole~K Speer, Khena~M Swallow, Todd~S Braver, and Jeremy~R Reynolds.
\newblock Event perception: a mind-brain perspective.
\newblock \emph{Psychological bulletin}, 133\penalty0 (2):\penalty0 273, 2007.

\bibitem[Zhang et~al.(2023{\natexlab{a}})Zhang, Li, and Bing]{videollama}
Hang Zhang, Xin Li, and Lidong Bing.
\newblock Video-llama: An instruction-tuned audio-visual language model for video understanding.
\newblock \emph{arXiv preprint arXiv:2306.02858}, 2023{\natexlab{a}}.

\bibitem[Zhang et~al.(2023{\natexlab{b}})Zhang, Liu, Dong, Huang, Ling, Wang, Wang, and Qiao]{zhang2023movqa}
Hongjie Zhang, Yi Liu, Lu Dong, Yifei Huang, Zhen-Hua Ling, Yali Wang, Limin Wang, and Yu Qiao.
\newblock Movqa: A benchmark of versatile question-answering for long-form movie understanding.
\newblock \emph{arXiv preprint arXiv:2312.04817}, 2023{\natexlab{b}}.

\bibitem[Zhang et~al.(2024{\natexlab{a}})Zhang, Wang, Tang, Liu, Feng, Dai, and Jin]{zhang2024flash}
Haoji Zhang, Yiqin Wang, Yansong Tang, Yong Liu, Jiashi Feng, Jifeng Dai, and Xiaojie Jin.
\newblock Flash-vstream: Memory-based real-time understanding for long video streams.
\newblock \emph{arXiv preprint arXiv:2406.08085}, 2024{\natexlab{a}}.

\bibitem[Zhang et~al.(2024{\natexlab{b}})Zhang, Zhang, Li, Zeng, Yang, Zhang, Wang, Tan, Li, and Liu]{longva}
Peiyuan Zhang, Kaichen Zhang, Bo Li, Guangtao Zeng, Jingkang Yang, Yuanhan Zhang, Ziyue Wang, Haoran Tan, Chunyuan Li, and Ziwei Liu.
\newblock Long context transfer from language to vision.
\newblock \emph{arXiv preprint arXiv:2406.16852}, 2024{\natexlab{b}}.

\bibitem[Zhang et~al.(2024{\natexlab{c}})Zhang, Zhang, Li, Zeng, Yang, Zhang, Wang, Tan, Li, and Liu]{zhang2024longva}
Peiyuan Zhang, Kaichen Zhang, Bo Li, Guangtao Zeng, Jingkang Yang, Yuanhan Zhang, Ziyue Wang, Haoran Tan, Chunyuan Li, and Ziwei Liu.
\newblock Long context transfer from language to vision.
\newblock \emph{arXiv preprint arXiv:2406.16852}, 2024{\natexlab{c}}.

\bibitem[Zhang et~al.(2025)Zhang, Yang, Yin, Luo, and Luan]{zhang2025qframequeryawareframeselection}
Shaojie Zhang, Jiahui Yang, Jianqin Yin, Zhenbo Luo, and Jian Luan.
\newblock Q-frame: Query-aware frame selection and multi-resolution adaptation for video-llms, 2025.

\bibitem[Zhang et~al.(2024{\natexlab{d}})Zhang, Wu, Li, Li, Ma, Liu, and Li]{llavavideo}
Yuanhan Zhang, Jinming Wu, Wei Li, Bo Li, Zejun Ma, Ziwei Liu, and Chunyuan Li.
\newblock Video instruction tuning with synthetic data, 2024{\natexlab{d}}.

\bibitem[Zhou et~al.(2024)Zhou, Shu, Zhao, Wu, Xiao, Yang, Xiong, Zhang, Huang, and Liu]{zhou2024mlvu}
Junjie Zhou, Yan Shu, Bo Zhao, Boya Wu, Shitao Xiao, Xi Yang, Yongping Xiong, Bo Zhang, Tiejun Huang, and Zheng Liu.
\newblock Mlvu: A comprehensive benchmark for multi-task long video understanding.
\newblock \emph{arXiv e-prints}, pages arXiv--2406, 2024.

\bibitem[Zhu et~al.(2025)Zhu, Wang, Chen, Liu, Ye, Gu, Duan, Tian, Su, Shao, et~al.]{zhu2025internvl3}
Jinguo Zhu, Weiyun Wang, Zhe Chen, Zhaoyang Liu, Shenglong Ye, Lixin Gu, Yuchen Duan, Hao Tian, Weijie Su, Jie Shao, et~al.
\newblock Internvl3: Exploring advanced training and test-time recipes for open-source multimodal models.
\newblock \emph{arXiv preprint arXiv:2504.10479}, 2025.

\end{thebibliography}
}


\appendix
\maketitlesupplementary

\section{SceneQA Further Analysis}

\begin{figure}[h]
    \centering
    \includegraphics[width=1.0\linewidth]{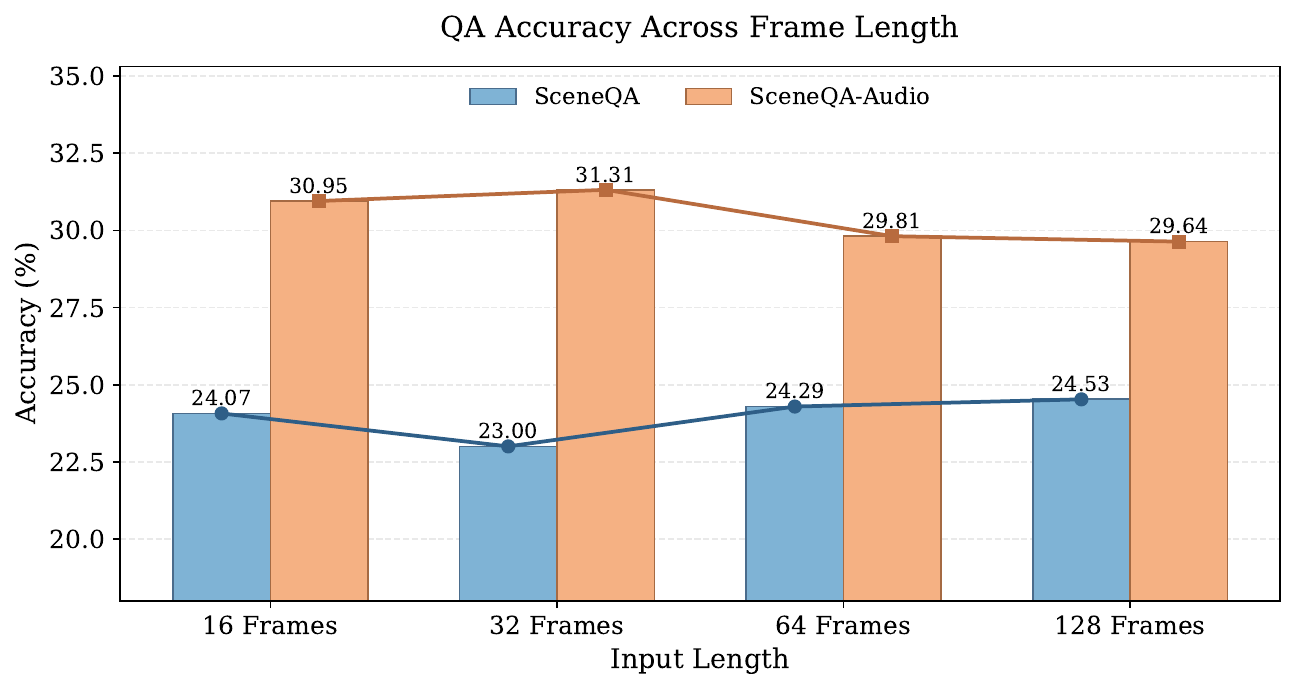}
    \caption{\textbf{QA Accuracy Across Frame Lengths}. Performance of SceneQA and SceneQA-Audio across different input frame lengths (16, 32, 64, and 128 frames). While SceneQA shows slight improvement with longer inputs, SceneQA-Audio performance peaks at moderate frame lengths and slightly declines for longer sequences.}
    \label{fig:qa1}
\end{figure}

Figure \ref{fig:qa1} analyzes the impact of input frame length on question answering performance. SceneQA shows a slight improvement as the number of frames increases, indicating that visual-only models benefit from longer temporal context. In contrast, SceneQA-Audio achieves its best performance at 32 frames, with accuracy gradually declining as the frame length increases, suggesting that longer input sequences may introduce noise or redundant information in the audio modality. This trend indicates that while extended visual context can be beneficial, incorporating audio signals reduces the need for longer inputs and may even be negatively affected by excessively long temporal context.

\begin{table}[t]
\centering
\caption{Temporal span distribution of SceneQA and SceneQA-Audio. 
We report the number of samples and average temporal span (seconds) for each temporal range.}
\label{tab:temporal_span}
\resizebox{0.47\textwidth}{!}{
\begin{tabular}{l|cc|cc}
\toprule
\rowcolor{headerpurple}
\textbf{Temporal} &
\multicolumn{2}{c|}{\textbf{SceneQA}} &
\multicolumn{2}{c}{\textbf{SceneQA-Audio}} \\

\rowcolor{headerpurple}
\textbf{Range (s)} &
\makecell{\textbf{\#}\textbf{Samples}} &
\makecell{\textbf{Avg.} \textbf{Span}} &
\makecell{\textbf{\#}\textbf{Samples}} &
\makecell{\textbf{Avg.}\textbf{Span}} \\

\midrule
$[0, 250)$         & 2110 & 163.13  & 2953 & 158.95 \\
\rowcolor{headerpurple} $[250, 500)$       & 700  & 342.91  & 769  & 337.94 \\
$[500, 750)$       & 256  & 608.29  & 187  & 588.53 \\
\rowcolor{headerpurple} $[750, 1000)$      & 57   & 853.72  & 34   & 861.50 \\
$[1000, 1250)$     & 68   & 1139.16 & 22   & 1109.36 \\
\rowcolor{headerpurple}$[1250, 1500)$     & 16   & 1356.12 & 29   & 1344.28 \\
$[1500, 1750)$     & 9    & 1582.33 & 4    & 1641.75 \\
\rowcolor{headerpurple}$[1750, 2000)$     & 15   & 1895.07 & 9    & 1883.56 \\
$[2000, +\infty)$  & 56   & 2453.14 & 10   & 2270.70 \\

\bottomrule
\end{tabular}
}
\end{table}

Table \ref{tab:temporal_span} summarizes the temporal span distribution of SceneQA and SceneQA-Audio in our dataset. The timespan is calculated based on the duration of context required by models to answer each question, averaged across models. Both subsets cover a wide range of temporal distances, from short-range events under 250 seconds to long-range dependencies exceeding 2000 seconds. Most samples are concentrated in shorter temporal ranges (below 500 seconds), following a natural distribution, while we intentionally include longer-context questions to encourage long-range reasoning. Importantly, SceneQA and SceneQA-Audio exhibit comparable average spans within each bucket, ensuring consistent temporal coverage across subsets.


\section{Runtime Latency Analysis}

We report the runtime of Scene-RAG breaking it down into offline preprocessing (video embedding, scene tiling, and audio captioning) and online inference (query rewriting and retrieval of relevant scene/audio context) in Table~\ref{tab:latency_breakdown}. 
A direct same-compute comparison is fundamentally flawed for long videos, since feeding equivalent frames directly causes out-of-memory while processing frames sequentially forces the baseline to treat the long video as disjointed short clips, destroying the global context required for reasoning.

\begin{table}[h]
    \centering
    \captionsetup{font=small}
    \caption{Runtime latency breakdown (video Length: 2,767s).}
    \vspace{-0.1in}
    \label{tab:latency_breakdown}
    \renewcommand{\arraystretch}{0.9}
    \resizebox{0.95\linewidth}{!}{
        \begin{tabular}{l|ccc|c}
            \toprule
            {Stage} & {Visual Enc.} & {Audio} & {LLM} & {Total} \\
            & \scriptsize{(InternVideo2)} & \scriptsize{(QwenAudio2)} & \scriptsize{(Qwen3)} & \scriptsize{(Seconds)} \\
            \midrule
            {Offline Preprocess} & 273.52 & 3.20 & - & \textbf{276.72} \\
            {Online Inference}   & 1.04 & 0.19 & 61.06 & \textbf{62.29} \\
            \bottomrule
        \end{tabular}
    }
    \vspace{-10pt}
\end{table}

\section{Annotation Details with example}
\begin{figure*}[t]
  \centering
  \vspace{-0.1in} 
  \includegraphics[width=0.98\textwidth]{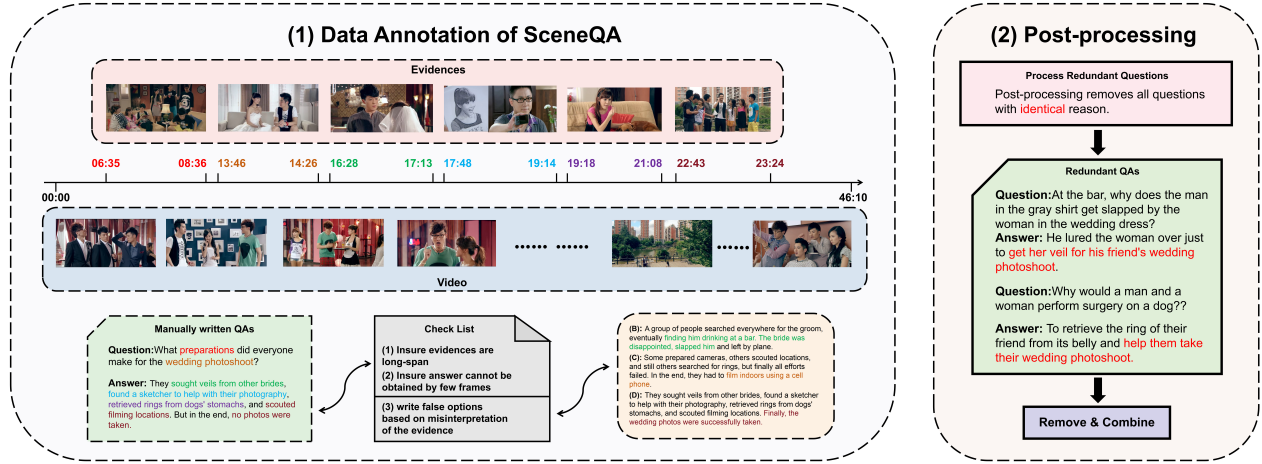}
  \vspace{-0.1in} 
  \caption{Overview of the Annotation pipeline. Zoom in for better visibility. }
  \label{fig:pipeline}
  \vspace{-5pt} 
\end{figure*}

Our benchmark is entirely manually annotated, specifically, Fig.~\ref{fig:pipeline} depicts the detailed annotation process for SceneQA. In terms of the Distance Definition, it is defined as the cue's observable interval. Regarding reproducibility, we will make our code and data publicly available before the deadline of camera ready.

\section{Related Work: RAG for Long Video}
Recent efforts to integrate Retrieval-Augmented Generation (RAG) with Multimodal Large Language Models (MLLMs) for long video understanding can be broadly categorized into online and offline approaches. 
In this work, we focus on offline RAG, where visual features are pre-extracted and reused for efficient query-based retrieval.  

VideoRAG~\cite{jeong2025videoragretrievalaugmentedgenerationvideo} adopts a straightforward pipeline: frames are sampled, and stored in a vector database for similarity search. While effective, uniform frame sampling can cause significant information loss.
Video-RAG~\cite{luo2024videoragvisuallyalignedretrievalaugmentedlong} enhances alignment by jointly modeling visual and textual semantics for more coherent retrieval. However, its dense representation leads to high memory usage and slower retrieval.
Q-Frame~\cite{zhang2025qframequeryawareframeselection} improves efficiency through query-aware frame selection, reducing redundant storage while maintaining relevance. Yet, it still requires handling large-scale video data.
To address this, MemVid~\cite{yuan2025memoryenhancedretrievalaugmentationlong} proposes a memory-enhanced framework that organizes features into hierarchical memory slots, enabling more compact and context-aware retrieval for long videos.

Although MemVid improves efficiency in memory construction and retrieval, it assumes videos are composed of continuous clips. In contrast, our work targets scene-based reasoning, where semantically related content can be discontinuous and scattered across different segments of the video, posing new challenges beyond existing long video RAG methods.

\section{Implementation Details of Scene-RAG}

We implement Scene-RAG using PyTorch. For visual representation, we utilize the pre-trained InternVideo2-6B~\cite{wang2024internvideo2} backbone, frozen during inference. Audio streams are processed using Qwen-Audio2~\cite{chu2024qwen2} to extract captions for speech and background sound. For the Large Language Model (LLM) backbone, we employ Qwen3-14B~\cite{yang2025qwen3}. The algorithm steps are summarized in Alg.~\ref{alg:pipeline}.

\paragraph{Hyperparameters.} The TV-L1 smoothing utilizes a regularization weight $\mu=0.5$. The sensitivity parameter for scene detection is set to $\alpha=1.5$ based on validation set performance. We filter out short segments with duration $L_{\min} < 3.0s$ to minimize noise. 
For retrieval, we maintain a memory bank size dynamic to the video length, retrieving the top-$K$ ($K=10$) most relevant scenes for final generation.

\paragraph{Ablation Study.}
We conduct a controlled ablation to isolate the contributions of (i) adaptive scene-sensitivity thresholding,
$k = \mu_x + \alpha\sigma_x$, (ii) TV-L1 smoothing for scene-tiling continuity (with $\alpha=1.5$), and (iii) the number of retrieved scenes $K$ used during memory-bank retrieval.
First, removing the adaptive thresholding and using a fixed cutoff leads to a notable drop in scene-boundary accuracy, confirming that dynamic scaling with $\alpha$ better adapts to local motion statistics.
Second, disabling TV-L1 smoothing (with default $L_{\min}=3$) causes fragmented scene boundaries and increases false splits, demonstrating the importance of regularized temporal gradients for stable segmentation.
Finally, we vary the top-$K$ retrieved scenes (with default $K=10$) and observe that too small a value underutilizes contextual history, while overly large $K$ introduces irrelevant or noisy scenes. 
Table~\ref{tab:ablation_scene_params} summarizes these findings, where we evaluated a grid search of hyperparameters.  We did not exhaustively explore all settings due to the computational cost of the experiments.

\begin{table}[t]
\centering
\caption{Ablation over SceneTiling $(\alpha,L_{\min})$ and Scene-RAG retrieval size $K$ on VideoMME \cite{videomme}. 
Model: Longva \cite{longva}.}
\label{tab:ablation_scene_params}
\resizebox{0.47\textwidth}{!}{
\begin{tabular}{l|ccc|cc}
\toprule
\rowcolor{headerpurple}
\textbf{Setting} &
\makecell{\textbf{TV-L1}\\$\alpha$} &
\makecell{\textbf{TV-L1}\\$L_{\min}$(s)} &
\makecell{\textbf{Scene-RAG}\\$K$} &
\makecell{\textbf{Avg.}\\\textbf{Result}} &
\makecell{\textbf{Gain}} \\

\midrule
Full Model (Ours) 
& 1.5 & 3.0 & 10 & 62.4 & - \\

\midrule
\rowcolor{headerpurple}$\alpha$ ↓ (0.5) 
& 0.5 & 3.0 & 10 & 61.2 & -1.2 \\

$\alpha$ ↑ (2.0) 
& 2.0 & 3.0 & 10 & 61.7 & -0.7 \\

\rowcolor{headerpurple}$L_{\min}$ ↓ (2s) 
& 1.5 & 2.0 & 10 & 61.9 & -0.5 \\

$K$ ↓ (5)
& 1.5 & 3.0 & 5 & 62.1 & -0.3 \\

\rowcolor{headerpurple}$K$ ↑ (15)
& 1.5 & 3.0 & 15 & 62.0 & -0.4 \\

\bottomrule
\end{tabular}
}
\end{table}


\section{Labeling Challenges.}  
Annotating long and complex videos presents significant challenges, especially for the SceneQA and I-VQA tasks. Both questions and answers can be ambiguous. For instance, a reference to “the person in red” may correspond to multiple individuals appearing at different times, while visually similar scenes, such as different classrooms, can cause confusion when identifying the correct location. These ambiguities require precise temporal localization and careful contextual verification. Annotators often need to re-watch the entire video, confirm spatial and temporal references, and cross-check with others to ensure that each question–answer pair aligns with a single, unambiguous narrative. On average, annotating one QA pair takes about 36 minutes, excluding the additional time required for review and verification. If a scene cannot be reliably characterized due to visual blurring, semantic ambiguity, or unclear narrative boundaries, we will directly abandon annotation for that scene. This ensures that all scene-level questions in SceneBench possess clear semantic support and verifiability.

\section{Ethical Concern and Data Publication}
\label{sec:ethical}

We do not own the video data. Instead, we collect access to publicly available videos in accordance with their original licensing conditions. We make reasonable efforts to ensure that the collected videos are legally redistributable and do not contain privacy-sensitive, illegal, or otherwise inappropriate content. The dataset will be released on HuggingFace.

\section{Supplementary Task Examples}

\begin{figure*}[p]
    \centering
    \includegraphics[width=0.95\textwidth]{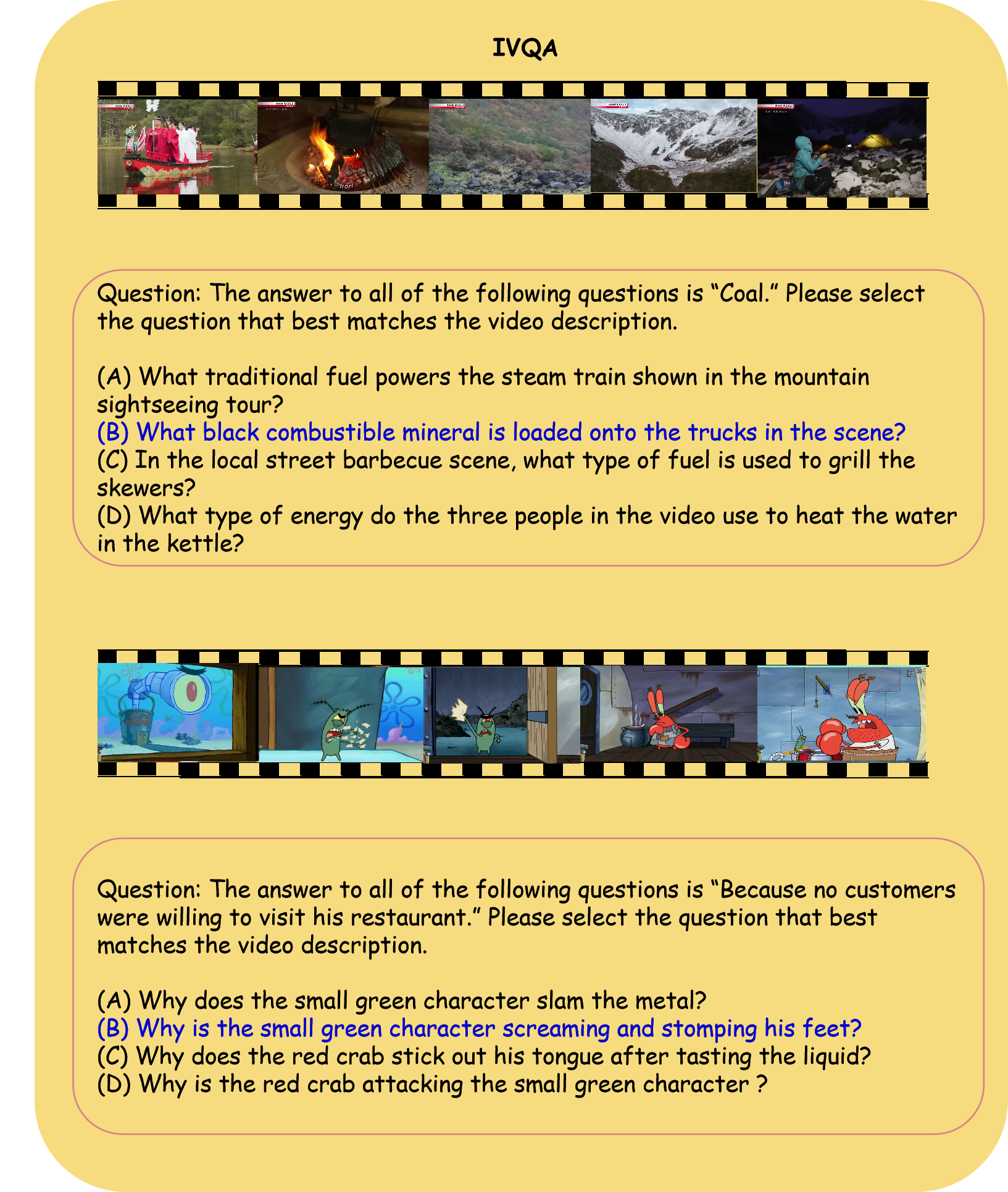}
    \caption{Examples QAs of \ourdataset. Overview of the problem set.}
    \label{fig:benchmark_detais}
\end{figure*}

\begin{figure*}[h]
    \centering
    \includegraphics[width=0.95\textwidth]{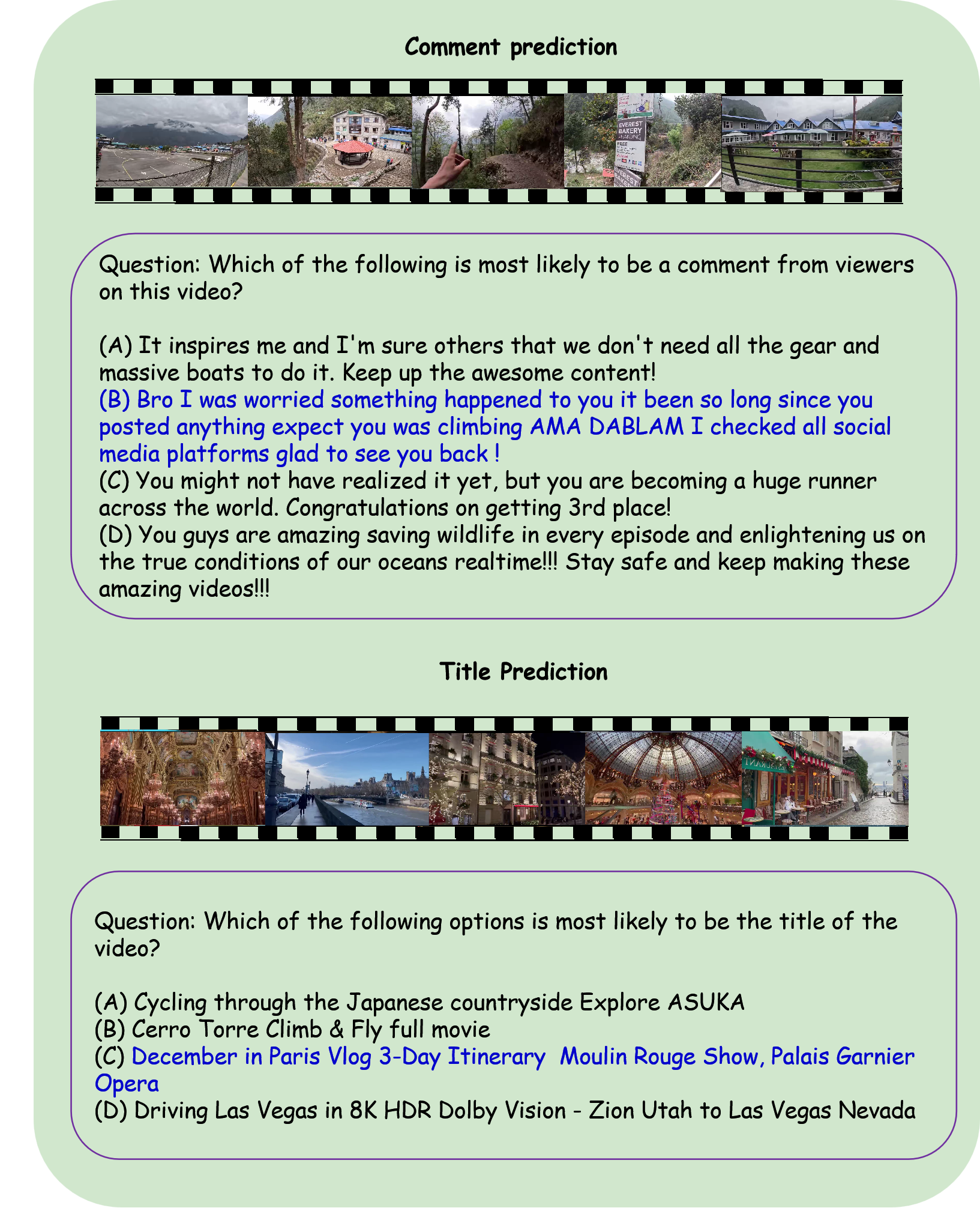}
    \caption{Examples QAs of \ourdataset. Overview of the problem set.}
    \label{fig:benchmark_detais2}
\end{figure*}

We also provide examples of I-VQA, Comment Prediction, and Title Prediction in Figure~\ref{fig:benchmark_detais} and Figure~\ref{fig:benchmark_detais2}.




\end{document}